\documentclass[10pt,twocolumn,letterpaper]{article}

\usepackage[pagenumbers]{cvpr}      
\usepackage{graphicx}
\usepackage{booktabs}
\usepackage{multirow}
\usepackage{amsmath,amssymb}
\usepackage{xcolor}
\usepackage{cuted}
\usepackage{caption}
\usepackage{float}
\usepackage{microtype}
\usepackage{xspace}
\newcommand{\model}{BADAS-3.0\xspace}

\renewcommand{\topfraction}{0.95}
\renewcommand{\bottomfraction}{0.6}
\renewcommand{\textfraction}{0.05}
\renewcommand{\floatpagefraction}{0.85}
\renewcommand{\dbltopfraction}{0.95}
\renewcommand{\dblfloatpagefraction}{0.85}
\definecolor{cvprblue}{rgb}{0.21,0.49,0.74}
\usepackage[breaklinks,colorlinks,allcolors=cvprblue]{hyperref}

\def\confName{CVPR}
\def\confYear{2027}

\title{BADAS-3.0: Real-Time Driving Perception and Planning from One Camera}
\hypersetup{pdftitle={BADAS-3.0: Real-Time Driving Perception and Planning from One Camera}, pdfauthor={Roni Goldshmidt, Hamish Scott, Lorenzo Niccolini}}

\author{Roni Goldshmidt \qquad Hamish Scott \qquad Lorenzo Niccolini\\
Empiric Earth\\
{\tt\small \{roni.goldshmidt, hamish.scott, lorenzo.niccolini\}@empiricearth.com}
}

\begin{document}
\maketitle
\begin{abstract}
We ask how far a driving model can get with one camera and real-time latency. \model is a 1B-parameter video encoder with four small heads. From two seconds of one forward camera, it predicts the current speed, the metric path just driven, a collision-risk score and six candidate futures over 6.4 seconds, in under 50 ms. Given the car's current speed, \model has 44\% lower minADE$_6$ over five trajectory benchmarks than the 34B Alpamayo~2~Super, which reads six cameras and runs over 100$\times$ slower. The gain comes from camera rigs that neither model was trained on. On the rig they share, the two are on par. Existing benchmarks are mostly nominal driving, so we add two open-loop safety benchmarks, one of real dashcam collisions and near-misses and one of Euro NCAP scenarios. On both, \model has a higher pass rate than every published planner we tested.
\end{abstract}

\section{Introduction}
\label{sec:intro}

\begin{figure*}[!t]
  \centering
  \includegraphics[width=\textwidth]{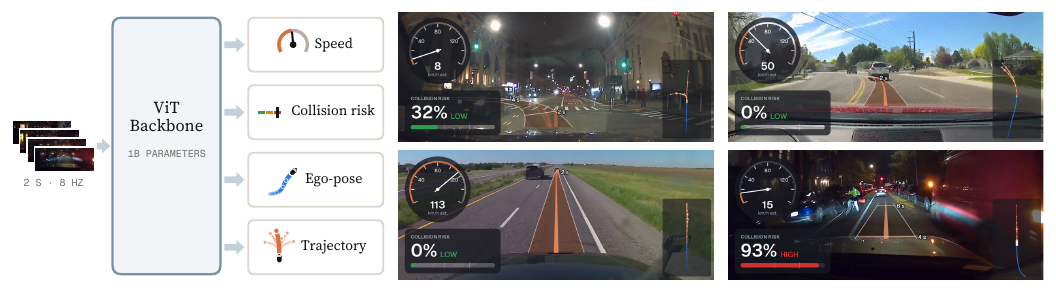}
  \caption{\textbf{\model.} A ViT backbone encodes two seconds of footage from one camera and four output heads return the speed, the past ego-pose, the collision risk and six candidate future trajectories.}
  \label{fig:teaser}
\end{figure*}

Driving planning models keep getting larger. Recently published systems are typically vision-language-action (VLA) models with several to tens of billions of parameters that read multiple cameras plus ego state or history~\cite{alpamayor1,alpamayocards,autovla2025,recogdrive2026,qwendrive}. These models may use long reasoning traces and diffusion or flow matching heads which require a large amount of iterative computation to produce driving trajectories. For example the recently released Alpamayo~2~Super takes 5 s and nearly 80 GiB of memory per two-second input window on an A100 GPU (Table~\ref{tab:runtime}, supplement).

We introduce \model, a 1B-parameter video vision transformer (ViT)~\cite{vit} taking as input a two-second clip from a single forward-facing camera. The model has four output heads: a trajectory head producing six candidate future trajectories as well as a soft selection score, a speed head regressing the current ego speed, an ego-pose head which predicts the metric path just driven and a risk head predicting the future collision-risk score (Figure~\ref{fig:teaser}). A single forward pass takes 43 ms and 1.9 GiB of memory (compiled, single A100, batch one).

We evaluate \model's trajectories on five driving benchmarks against several open driving models (Section~\ref{sec:protocol}). Given the car's current speed, \model has 44\% lower minADE$_6$ than Alpamayo~2~Super, which has 34$\times$ more parameters, reads six cameras and runs over 100$\times$ slower (Section~\ref{sec:traj}). Adding the rest of the ego history improves this by less than 1\%. With no speed input at all, \model uses its own speed estimate. Calibrating that estimate on ten examples recovers most of the lost accuracy. \model's mean error is then below every baseline's, even though the baselines get the full ego history.

Existing benchmarks are mostly nominal driving, and predicting what the driver would do is not the same as avoiding a collision. We therefore add two open-loop safety benchmarks that test whether the model's trajectory avoids a collision in a high-risk event. The Euro NCAP benchmark replays the protocol's collision-avoidance scenarios in simulation, and the Dashcam Collision Avoidance benchmark (DCA-Bench) uses real-world collisions and near-misses from dashcam footage. \model has a higher pass rate than any published planner we tested.

Finally, we evaluate \model on the planning benchmark NavSim~\cite{navsim2024}. With the base model frozen, we fit six calibration parameters and train a selector for this benchmark. \model then scores 89.0 PDMS, and 91.1 after fine-tuning (Section~\ref{sec:navsim}).

We also evaluate the speed, ego-pose and collision-risk heads, each against other systems for that task. For speed and ego-pose we compare against visual odometry and 3D reconstruction systems, and \model has the lowest mean error. The model takes no camera intrinsics. The evaluation uses a camera's focal length only to frame the image to the training field of view (Section~\ref{sec:traj}). On a camera outside the training range, \model's speed and ego-pose are off by a roughly constant factor. One scale per camera, fitted from ten labelled examples, removes most of that error. On five collision anticipation benchmarks, the risk head is within 0.02 AP of its teacher, the state-of-the-art BADAS-2.0~\cite{badas2}.

We make three contributions. First, a 1B single-camera model with the best mean accuracy on trajectory (with a speed sensor), ego-pose, speed and collision-risk benchmarks, at up to 100$\times$ lower latency than frontier VLAs. Second, metric ego-pose and speed from one camera with no intrinsics given to the model, and a ten-example scale correction for cameras outside its training range. Third, two open safety benchmarks on which existing planners fail most scenarios.\looseness=-1

We will release \model open-source, along with the five benchmark files, the evaluation harness and our open-loop safety benchmarks.

\section{Related Work}
\label{sec:related}

\textbf{Self-supervised video representations.} Generative world models such as GAIA-1 and GAIA-2~\cite{gaia1,gaia2} and the Cosmos family~\cite{cosmos1,cosmostransfer,cosmos3} synthesise driving video from actions and scene attributes. \model instead builds on V-JEPA~\cite{vjepa}, which learns by predicting masked regions in latent space and showed that frozen probes on its representations rival task-specific systems; V-JEPA~2~\cite{vjepa2} and V-JEPA~2.1~\cite{vjepa21} scale this to 1B and 2B parameters respectively. Other large video encoders include VideoMAE~V2~\cite{videomaev2}, which reconstructs pixels at a similar scale, and Cosmos-Embed1~\cite{cosmosembed1}, an industrial video embedder. Drive-JEPA~\cite{drivejepa} also pretrains a V-JEPA encoder on driving video, for a NavSim planner on one front camera; \model adds metric speed, ego-pose and collision-risk heads and is evaluated across camera rigs. We find that the V-JEPA objective trained in-domain on dashcam video outperforms general video data (Section~\ref{sec:backbone}, supplement), and use such a model to initialise the backbone of \model.

\textbf{Vision-language-action policies} such as Alpamayo-R1~\cite{alpamayor1} pair a Cosmos-Reason VLM with a diffusion-based action expert over the future horizon, with chain-of-causation reasoning trained by supervised fine-tuning and reinforcement learning. Its successors scale the same recipe, to 10.5B parameters for Alpamayo~1.5 and 34B for Alpamayo~2~Super, which reads six cameras~\cite{alpamayocards}. Qwen-Drive~\cite{qwendrive} adds a planning expert to a 4B multimodal VLM that also handles 3D perception and scene question answering, and SimWAM~\cite{simwam} co-trains a pretrained video expert with a lightweight action expert and discards the video branch at inference. These planners take seconds per decision (Table~\ref{tab:runtime}, supplement). Fixed anchors with learned offsets go back to MultiPath~\cite{multipath}, which predicts a distribution over a fixed set of anchor trajectories and regresses offsets for each. We use the same idea in control space, with anchors that are sequences of acceleration and curvature.

\textbf{Monocular ego-motion.} From one camera, metric scale is unobservable without extra sensing or priors, and methods differ in how they recover it. XVO~\cite{xvo} regresses metric motion from pixels; ZeroVO~\cite{zerovo}, OpenVO~\cite{openvo} and UniDepth-V2 with PnP~\cite{unidepth} estimate intrinsics or depth explicitly; VGGT~\cite{vggt}, DGGT~\cite{dggt}, MASt3R-SLAM~\cite{mast3rslam} and DPVO~\cite{dpvo} recover paths up to scale~\cite{dust3r,mast3r}. Our head is metric with no intrinsics, no alignment, causal and stateless.

\textbf{Collision anticipation.} Anticipating a crash from one forward camera was posed as frame-level prediction on DAD~\cite{dad} and later benchmarks~\cite{ustring,dada2000,dota,nexardataset}. Early models attend over detected objects~\cite{dsta,ustring,drive}; later ones add context or robustness~\cite{crash,accnet,predictresist}, language and attention supervision~\cite{cap,mmau,w3al}, or new labels~\cite{top,riskprop}, and video foundation encoders now lead the benchmarks~\cite{vagnet,videomaezs,orlova2025tad,movad}. The BADAS series, BADAS-1.0~\cite{badas1} and BADAS-2.0~\cite{badas2}, achieved state-of-the-art performance by fine-tuning V-JEPA~2~\cite{vjepa2} for collision anticipation but these models were specialised to only this task. \model uses BADAS-2.0 as the risk head's teacher but expands beyond just collision anticipation. 

\textbf{Safety benchmarks.} Trajectory benchmarks such as nuScenes~\cite{nuscenes}, Argoverse~2~\cite{av2} and Waymo~\cite{wod} score a planner against the path the driver took, and those recordings are almost entirely safe, routine driving. NavSim~\cite{navsim2024} adds no-collision and time-to-collision terms to its score on recorded drives. AlpaSim~\cite{alpamayor1} replays 75 recorded drives chosen for dense interactions and counts the close encounters a planner causes. WOD-E2E~\cite{wode2e} rates plans on long-tail scenes against human preference. In each, the recorded outcome is safe and the planner fails by leaving it. NeuroNCAP~\cite{neuroncap}, Bench2Drive~\cite{bench2drive} and SafeBench~\cite{safebench} do place hazards on a collision course, in closed loop, on synthetic or reconstructed scenes. What is missing is a test that starts from a collision already under way, on a real road as well as on a protocol, and that any released planner can take open-loop, from the same frames, without being wired into a simulator.

\section{The Model}
\label{sec:model}

\model consists of a 1B-parameter ViT backbone with four separate heads. Each head is an attentive pooler followed by a small network.

\textbf{Backbone.}
The backbone takes as input a single 2 s video clip at $256\times 256$ resolution and 8 fps. It is initialised from a model pretrained from scratch with the V-JEPA objective~\cite{vjepa} over around 110,000 hours of proprietary dashcam video. All four heads are lightweight readouts of the same backbone features. The backbone uses a patch size of $16\times16\times2$ ($H\times W\times T$), so the input clip becomes 2048 tokens of 1408 dimensions each. Forty transformer layers turn these tokens into the output patch features.\looseness=-1

\subsection{Trajectories via controls}
\label{sec:heads}

To output smooth driving trajectories \model produces a sequence of acceleration and curvature controls, $a_t, \kappa_t$. Then a unicycle model converts these into a smooth trajectory given an initial speed, $v_0$: 

\begin{equation}
\small\begin{aligned}
v_{t+1}&=\max(0,\,v_t+a_t\,\Delta t), & \psi_{t+1}&=\psi_t+v_t\kappa_t\,\Delta t,\\
x_{t+1}&=x_t + v_t\,\Delta t\,\cos \psi_t, & y_{t+1}&=y_t + v_t\,\Delta t\, \sin\psi_t.
\end{aligned}
\label{eq:unicycle}
\end{equation}

We use $\Delta t=0.1$s and roll out trajectories for 64 steps giving future trajectories up to 6.4s ahead. The $\max$ in the speed update floors the speed at zero, so a candidate that brakes to a stop stays stopped instead of rolling backward. Training and the safety selector use the update without this floor, so the floor changes only the geometry of candidates that stop. Table~\ref{tab:nofloor} (supplement) measures its effect. The one boundary condition is $v_0$, which comes either from an external speed sensor or from \model's speed head. Section~\ref{sec:speedsources} defines the sensor, the zero-shot speed head and a speed head calibrated from ten examples, and Section~\ref{sec:traj} compares them and benchmarks the speed head itself.

\model produces $K=6$ candidate trajectories from each input rather than just one. To do this it uses a set of mode anchors, $\mathbf{A}_k \in \mathbb{R}^{64\times 2}$, which are fixed sequences of acceleration and curvature controls. These are precomputed via $k$-means over the training set (Figure~\ref{fig:anchors_xy}, supplement). The model outputs a set of mode deltas $\Delta_k \in \mathbb{R}^{64\times 2}$ which are added to their corresponding anchor to give the unicycle model inputs: $[a_t^{(k)}, \kappa_t^{(k)}] = \mathbf{A}_k + \Delta_k$.

The trajectory head is an attentive pooler. Six learned mode queries attend to the backbone's patch tokens together with an embedding of the ego history, which is replaced by learned null tokens when no history is given, and a linear layer turns each resulting hidden state $\mathbf{h}_k$ into the mode deltas $\Delta_k$ (architecture details in Section~\ref{sec:headarch}, supplement).

The head is trained with a closest-mode loss. Each of the $K$ control sequences is rolled out through Equation~\ref{eq:unicycle} to a path $\hat{\tau}^{(k)} \in \mathbb{R}^{64\times 2}$, and the mode whose path lies nearest the ground truth $\tau$ is selected, $k^\star = \arg\min_k \|\hat{\tau}^{(k)} - \tau\|$. Only that mode receives a regression gradient, a Huber loss with $\beta = 1$ m on its $xy$ waypoints, and the selection logits $\mathbf{s} \in \mathbb{R}^K$ are trained with a cross-entropy loss to identify it:
\begin{equation}
\mathcal{L}_\text{trajectory} = \text{Huber}\big(\hat{\tau}^{(k^\star)}, \tau\big) + 0.5\,\text{CE}\big(\mathbf{s}, k^\star\big).
\label{eq:trajloss}
\end{equation}
Supervising the integrated path rather than the controls trains the head in the space the metrics measure. Since only the closest mode is pulled onto the truth, braking scenes train the braking anchors and cruising scenes the cruising ones, which keeps the six candidates diverse (Figure~\ref{fig:anchors}, supplement).

\textbf{Safety selector.} The logits $\mathbf{s}$ are trained only to identify $k^\star$, the candidate closest to the driven path, and nothing in that rewards safety: a driver who crashed is matched as closely as one who braked. \model therefore adds a second ranking over the same six candidates, trained to prefer the safer one when the scene carries risk and to agree with the driver otherwise. The safety selector scores each candidate from $h_k$, its rolled-out path and the logit of the risk head, and outputs a correction to $\mathbf{s}$. The car drives the candidate with the highest corrected score.

\subsection{Ego-pose, speed and risk}

\textbf{Ego-pose.} The ego-pose head predicts the metric path driven by the ego over the 2 s of input video. It applies attentive pooling to spatially pooled patch features and outputs four parameters $[a_0, a_1, \kappa_0, \kappa_1]$, which define the acceleration and curvature over the predicted window by:

\begin{equation}
\begin{aligned}
  a(t)&=a_0 + a_1 t & \kappa (t)&=\kappa_0 + \kappa_1 t, \\
\end{aligned}
\label{eq:egomotion}
\end{equation}
where $t$ is rescaled to run from -1 to 1 over the predicted window.

We then apply the unicycle model used for the future trajectory head but in reverse to compute the past ego trajectory.

\textbf{Speed.}
The speed head uses the same attentive pooling as the ego-pose head, with a single scalar output. This scalar is interpreted as $\log(\hat{v} + 0.1)$ where $\hat{v}$ is the predicted speed and is trained with a Huber loss against the ground truth, $\log(v + 0.1)$.

\textbf{Risk.}
The risk head applies attentive pooling to all patch tokens and an MLP to give the output logit. This is trained via distillation with a cross-entropy loss against the output of a teacher model which is BADAS-2.0~\cite{badas2}.

\subsection{Datasets}

For the driving data we use the training split of NVIDIA PhysicalAI-Autonomous-Vehicles~\cite{physicalai}, which we denote $\mathcal{D}_\text{PAI}$. It has 2.86M windows from about 239k clips, each pairing a two-second video clip with the ego's poses over the preceding 2 s and the following 6.4 s at 10 Hz. Our evaluation clips are removed.

We also use proprietary dashcam video annotated with ground-truth ego trajectories, which we recover from the video by 3D reconstruction and scale with GPS. We combine 87k of these windows with an equal number sampled from $\mathcal{D}_\text{PAI}$, and denote the mixture $\mathcal{D}_\text{mixed}$.

For collision risk we use 1.88M proprietary dashcam windows labelled by the BADAS-2.0 teacher~\cite{badas2}, which we denote $\mathcal{D}_\text{risk}$. Each window has a binary collision label and the teacher's collision probability, and no ego trajectory. All training data, including backbone pretraining, is PhysicalAI-AV or proprietary dashcam video.\looseness=-1

\subsection{Training}
\label{sec:training}
\model is trained in stages, because each head needs different data. 

\textbf{Initial training of backbone and trajectory head.}
We take the V-JEPA pretrained backbone and initialise the trajectory head. We train both end-to-end on $\mathcal{D}_\text{PAI}$ with the closest-mode loss of Equation~\ref{eq:trajloss} against the ground truth future trajectory. The backbone is frozen for the first 2,000 steps to allow the trajectory head to warm up before being unfrozen for a further 140,000 steps. Ego history is always provided in full.

Next we adapt the model to other camera configurations and to inputs without ego history. We train on $\mathcal{D}_\text{mixed}$ with the same loss, but drop the ego history half of the time and replace it with the learned null tokens. We update the head only for 500 steps and then the full model for another 8,500. We select the best checkpoint using validation ADE against a held-out set. 

\textbf{Adding the risk head.}
We now jointly train the model for trajectory forecasting and collision risk prediction. At each training step we sample two batches, one from $\mathcal{D}_\text{mixed}$ of clips and trajectories, $(c_i, \tau_i)$, and one from $\mathcal{D}_\text{risk}$ of clips and ground truth collision labels and risk probabilities from the BADAS-2.0 teacher model, $(c_j, r_j, p_j)$.

From our model we obtain a predicted trajectory and a predicted risk score from each batch respectively:
$$\hat{\tau}_i := f_\theta(c_i) \quad \hat{p}_j := f_\theta(c_j).$$
We then compute a combined loss against the relevant labels for each prediction, with $\mathcal{L}_\text{trajectory}$ as in Equation~\ref{eq:trajloss}:

\begin{equation}
\begin{array}{@{}l@{\hspace{0.5em}}l@{}}
  \mathcal{L}_\text{risk} & {=}\ (1 - \lambda) \text{BCE}(\hat{p}_j, r_j) + \lambda \text{BCE}(\hat{p}_j, p_j),\\
  \mathcal{L} & {=}\ \mathcal{L}_\text{trajectory} + w \mathcal{L}_\text{risk}.
\end{array}
\label{eq:loss}
\end{equation}

We use $w = 5$ throughout. We train for 25,000 steps. The first 12,500 use $\lambda = 0.5$ which balances between the true label and the teacher. The second 12,500 use $\lambda = 0.9$, which focuses on distilling the knowledge in the teacher. The gradient from the risk loss is only propagated to the backbone after 500 steps.

After this joint training phase we freeze the backbone and trajectory head and do a final phase which further distills the teacher model into the risk head. This step uses only $\mathcal{L}_\text{risk}$ with $\lambda = 1$ and trains for 6,000 steps.

\textbf{Training the safety selector.} Last, we train the safety selector alone. The backbone and all other heads are frozen, so the six candidates are fixed and only the choice among them is trained. We train on 30,696 windows from about 9,100 clips annotated with ground truth trajectories for the ego as well as all third parties. Its training clips are disjoint from the DCA-Bench test split. We roll out each candidate and check whether it collides with a third party. The winner is the candidate closest to the ego ground truth that does not collide. The safety selector is trained to pick the winning trajectory with a cross-entropy loss.

\textbf{The speed and ego-pose heads.} Both the ego-pose and speed heads are trained on $\mathcal{D}_\text{mixed}$ with the backbone frozen. For ego-pose we use the Huber loss against 20 poses corresponding to the input 2 second clip. For speed we use the Huber loss against $\log(v + 0.1)$ where $v$ is the ego speed at the end of the 2 second input clip. Both heads are trained for 80,000 steps.

\subsection{Runtime}
Compiled, on one A100 at batch one, the backbone and trajectory head take 43 ms, not counting preprocessing. The other heads and the safety selector add under 3 ms. On the same GPU, Qwen-Drive and the larger Alpamayo models take between 967 ms and 4.9 s (Table~\ref{tab:runtime} and Figure~\ref{fig:runtime}, supplement).

\newenvironment{tight}{\begin{itemize}\setlength{\itemsep}{1pt}\setlength{\parsep}{0pt}\setlength{\topsep}{2pt}\setlength{\parskip}{0pt}}{\end{itemize}}

\section{Experiments}
\label{sec:protocol}
\label{sec:results}

\subsection{Three sources of speed}
\label{sec:speedsources}

The unicycle model needs an initial speed, and we test three sources for it. In the \textbf{sensor} setting, the true speed comes from an external sensor or the ego history. In the \textbf{zero-shot} setting, the true speed is unavailable and \model uses its own speed head. In the \textbf{few-shot} setting, we have ten examples from the target dataset or vehicle, annotated with the true speed. We fit one scale to the speed head on them, the ratio of the summed true speeds to the summed predicted speeds. We apply the scale only when it differs from one by more than twice its standard error and by more than 5\%. Otherwise we keep the zero-shot prediction. The ten examples are not scored (Section~\ref{sec:calibration}, supplement).

\subsection{Benchmarking the output heads}
\label{sec:traj}
\label{sec:speed}
\label{sec:ego}
\label{sec:risk}

\textbf{Datasets.} For trajectory forecasting we use five public driving sets: nuScenes~\cite{nuscenes}, Argoverse~2 Sensor and TbV~\cite{av2}, Waymo~\cite{wod} and PAI-644, a held-out split of 644 PhysicalAI-AV clips~\cite{physicalai}. The ground truth is the driven path, 64 waypoints at 10 Hz from 0.1 to 6.4 s in the ego frame at $t_0$. Ego-pose uses the same five sets plus KITTI~\cite{kitti}, with two windows per PAI-644 clip. Speed uses the same sets as ego-pose but replaces PAI-644 with PAI-100c, 100 other PhysicalAI-AV clips. Collision risk uses five collision anticipation datasets under the conventions of~\cite{badas2}: Nexar Public~\cite{nexardataset}, the long-tail set of~\cite{badas2}, DAD~\cite{dad}, DoTA~\cite{dota} and DADA-2000~\cite{dada2000}. DAD, DoTA and DADA-2000 use the ego-centric labelling from \cite{badas1}.

\textbf{Baselines.} For trajectories we compare against the Alpamayo family~\cite{alpamayocards,alpamayor1,alpamayoedge}. Qwen-Drive's released head~\cite{qwendrive} stops at 5 s, so we compare with it at that horizon in the supplement (Table~\ref{tab:traj5}). Each model receives its native camera and ego-history inputs. \model was trained on 16:9 cameras whose focal length is 0.48 to 0.60 of the frame width. The front cameras of nuScenes, Argoverse~2 and Waymo are narrower. Before the $256\times256$ resize, we place their frames on a 16:9 canvas on which the focal length is 0.60 of the canvas width, the value of the training dashcam. The focal lengths come from each dataset's calibration, and cameras inside the training range are untouched (Section~\ref{sec:framing}, supplement). For speed and ego-pose we compare against metric visual-odometry pipelines and scale-free 3D-reconstruction systems, all run at their native resolution. Each baseline's speed is the magnitude of its last-step translation. For collision risk we compare against the earlier BADAS models~\cite{badas1, badas2} and several public VLMs.

\textbf{Metrics.} Trajectories are scored by minADE$_6$~\cite{nuscenes,av2,womd}, the average displacement error in metres of the best of six candidates against the ground truth. Speed is measured by the mean absolute error (MAE) in m/s of the predicted speed at the end of the clip. For ego-pose we report the median absolute trajectory error (ATE), the RMS distance in metres over the twenty poses of the two-second path in the ego frame at $t_0$, with a median taken over all windows. The raw ATE is unaligned, so it includes scale error. We also report ATE after a per-window similarity fit of rotation, translation and scale, which leaves only the shape error on which scale-free methods are comparable. Collision risk is measured via mAP@TTA on the Nexar Public dataset~\cite{nexardataset} (the mean AP at 0.5, 1.0 and 1.5 s before the accident) and clip-level AP elsewhere.\looseness=-1

\begin{table*}[!t]
\centering\scriptsize\setlength{\tabcolsep}{3.5pt}
\resizebox{\textwidth}{!}{\begin{tabular}{@{}llrrrrrrrr@{}}
\toprule
model & input & params & nuSc & AV2-S & AV2-TbV & Waymo & PAI-644 & mean$_5$ & top-1 \\
\midrule
\textbf{BADAS-3.0}, speed head, zero-shot & 1 cam & 1B & 1.223 & 1.440 & 1.286 & 3.192 & 1.272 & 1.682 & 3.912 \\
\textbf{BADAS-3.0}, speed head, few-shot & 1 cam & 1B & 1.217 & 1.234 & 1.061 & 1.518 & 1.275 & 1.261 & 3.300 \\
\textbf{BADAS-3.0}, speed sensor & 1 cam & 1B & 0.896 & 1.006 & \textbf{0.942} & 1.000 & 0.929 & 0.955 & 2.719 \\
\textbf{BADAS-3.0}, speed sensor + ego history & 1 cam & 1B & \textbf{0.894} & \textbf{1.005} & 0.946 & \textbf{0.988} & \textbf{0.927} & \textbf{0.952} & 2.711 \\
\midrule
Alpamayo 2 Super~\cite{alpamayocards} & 6 cam & 34B & 2.347 & 1.977 & 1.774 & 1.414 & 0.957 & 1.694 & 3.423 \\
Alpamayo-R1~\cite{alpamayor1} & 4 cam & 10B & 2.138 & 2.159 & 2.075 & 1.692 & 1.031 & 1.819 & 3.741 \\
Alpamayo 1.5~\cite{alpamayocards} & 4 cam & 10.5B & 2.117 & 2.289 & 1.893 & 1.438 & 0.954 & 1.738 & 3.344 \\
Alpamayo 2 Edge~\cite{alpamayoedge} & 4 cam & 2B & 1.745 & 1.738 & 1.971 & 1.262 & 1.269$^{\dagger}$ & 1.597$^{\dagger}$ & -- \\
\bottomrule
\end{tabular}}
\caption{\textbf{Future trajectory, minADE$_6$ over a 6.4-second horizon, in metres} (best of six candidates). Mean$_5$ is the mean over the five columns. Alpamayo-R1, 1.5 and 2~Edge read four cameras and Alpamayo~2~Super reads six, each with 1.6 s of ego history. \model reads the front camera only, and gets ego history (2 s, 20 poses) only in its last row. Top-1 is the ADE of one trajectory, averaged over the five sets. For \model it is the head's own pick. Alpamayo does not rank its samples, so its top-1 is the mean of its six. A dash means not computed. $^\dagger$Scored on NVIDIA's anchor windows of the 644 clips, not on the windows of the other rows. Alpamayo uses one seed. Over three seeds, Alpamayo~1.5 on nuScenes moves $\pm$0.13 m in minADE$_6$ and $\pm$0.18 m in top-1 (supplement).}
\label{tab:traj}
\end{table*}

\begin{figure}[!t]
  \centering
  \includegraphics[width=0.85\columnwidth]{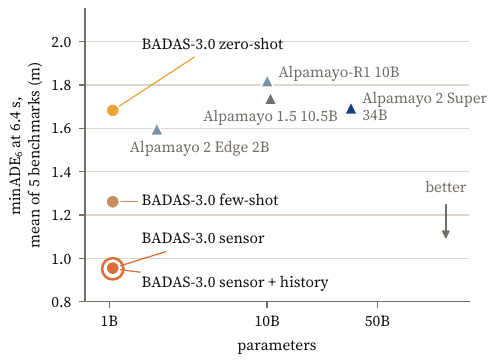}
  \caption{\textbf{Accuracy against size.} Mean minADE$_6$ over a 6.4-second horizon on the five test sets, against parameters, for the \model variants and Alpamayo family (Alpamayo~2~Edge: PAI-644 on other windows, Table~\ref{tab:traj}).}
  \label{fig:trajsize}
\end{figure}

\textbf{Trajectory forecasting.} With one camera, a speed sensor and the ego history, \model has a mean minADE$_6$ of 0.95 m (Table~\ref{tab:traj}, Figure~\ref{fig:trajsize}). That is 44\% lower than Alpamayo~2~Super at 1.69 m, which has 34$\times$ more parameters and reads six cameras. The lowest baseline mean is 1.60 m, from the 2B Alpamayo~2~Edge. \model is also ahead of Qwen-Drive at its 5 s horizon, at 0.64 m against 1.89 m (Table~\ref{tab:traj5}, supplement).

The gap comes from cameras the models were not trained on. Most of it is on nuScenes and the two Argoverse sets. On Waymo, \model is 0.27 m ahead of the best baseline. On PAI-644, which both \model and Alpamayo~2~Super were trained on, the two are essentially on par, at 0.93 m and 0.96 m.

Without a speed sensor, \model must use its own speed prediction, and the mean rises to 1.68 m (zero-shot). Calibrating the speed head on ten labelled examples recovers 58\% of that rise (few-shot, 1.26 m). The few-shot mean is below every baseline's mean.

minADE$_6$ credits the best of six candidates, but a car drives only one. \model's top-ranked candidate has a mean ADE of 2.71 m (top-1 column, Table~\ref{tab:traj}). Alpamayo~2~Super has 3.42 m, averaged over its six samples because it does not rank them. On PAI-644, Alpamayo's single paths are closer to the driver's. The safety selector's choice has a mean ADE of 2.73 m, closer than any Alpamayo model's single path (3.34 to 3.74 m).\looseness=-1

\begin{table*}[!t]
\centering\scriptsize\setlength{\tabcolsep}{3.5pt}
\resizebox{\textwidth}{!}{\begin{tabular}{@{}llrrrrrrrr@{}}
\toprule
method & scale & nuSc & AV2-S & AV2-TbV & Waymo & KITTI & PAI-644 & mean$_6$ & runtime \\
\midrule
\textbf{BADAS-3.0}, speed head, zero-shot & metric & 0.44 & 0.99 & 0.82 & 1.53 & 3.03 & \textbf{0.25} & 1.18 & 43 ms \\
\textbf{BADAS-3.0}, speed head, few-shot & metric & \textbf{0.43} & \textbf{0.40} & \textbf{0.31} & \textbf{0.47} & \textbf{0.53} & \textbf{0.25}$^{\ddagger}$ & \textbf{0.40}$^{\ddagger}$ & 43 ms \\
\midrule
XVO~\cite{xvo} & metric & 0.81 & 2.62 & 1.95 & 1.72 & 2.02 & 2.22 & 1.89 & 190 ms \\
ZeroVO~\cite{zerovo} & metric & 2.36 & 1.52 & 1.21 & 2.25 & 2.98 & 4.53 & 2.47 & 9.4 s \\
OpenVO~\cite{openvo} & metric & 3.76 & 3.40 & 1.88 & 2.05 & 3.56 & 3.60 & 3.04 & 5.8 s \\
UniDepth-V2 + PnP~\cite{unidepth} & metric & 3.97 & 2.04 & 1.61 & 0.86 & 1.19 & 2.13 & 1.97 & 1.0 s \\
\midrule
DPVO~\cite{dpvo} & scale-free & 6.30 & 7.09 & 4.92 & 6.21 & 9.43 & 8.39 & 7.06 & 667 ms \\
VGGT~\cite{vggt} & scale-free & 6.14 & 7.11 & 4.79 & 6.30 & 9.86 & 8.21 & 7.07 & 786 ms \\
DGGT~\cite{dggt} & scale-free & 6.26 & 7.15 & 4.80 & 6.31 & 9.89 & 8.21 & 7.10 & 1.1 s \\
MASt3R-SLAM~\cite{mast3rslam} & scale-free & 4.34 & 2.06 & 0.19 & 4.50 & 6.10 & 0.15 & 2.89 & 10.9 s \\
\bottomrule
\end{tabular}}
\caption{\textbf{Two-second ego-pose, median ATE in metres, no alignment.} Mean$_6$ is the mean over the six columns. Runtime is per window on one A100. The same windows after a per-window similarity fit are in the supplement (Table~\ref{tab:egosim3}). MASt3R-SLAM, scored only where it tracks, is unranked. Few-shot rows use Equation~\ref{eq:gate} of the supplement. $^{\ddagger}$Fitted with the earlier rule (Section~\ref{sec:calibration}).}
\label{tab:ego}
\end{table*}

\begin{figure}[t]
  \centering
  \includegraphics[width=0.9\columnwidth]{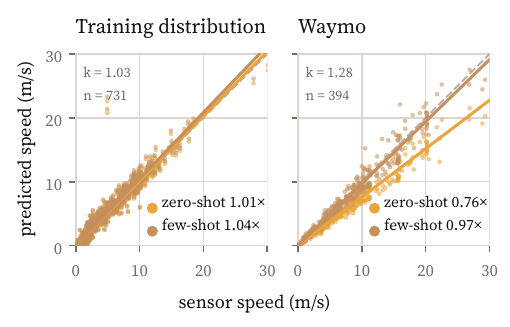}
  \caption{\textbf{One scalar moves the whole cloud.} Predicted against sensor speed per window. On Waymo, a camera the head never saw, zero-shot is a straight line at 0.76 of the truth (through-origin fit; the benchmark-wide scale bias is 0.75) and $k$ lifts it to 0.97. On the training distribution $k$ is already near one and the two clouds coincide.}
  \label{fig:speedscatter}
\end{figure}

\textbf{Speed estimation.} \model's speed head is more accurate than every odometry pipeline on five of six datasets and has the lowest mean MAE (Table~\ref{tab:speed}, supplement, and Figure~\ref{fig:speedscatter}). Calibrating on ten labelled examples cuts the mean error by 58\%. The largest gains are on Waymo (59\%) and KITTI (83\%).\looseness=-1

\textbf{Ego-pose prediction.} \model has the lowest mean error on both measures (Table~\ref{tab:ego}, and Table~\ref{tab:egosim3} and Figure~\ref{fig:egocost} in the supplement). Zero-shot, its Mean$_6$ ATE is 1.18 m. The metric visual-odometry pipelines are at 1.89 to 3.04 m and take 190 ms to 9.4 s per window, against 43 ms. After a per-window similarity fit removes scale, \model's Mean$_6$ error is 3.5 cm, and the next system, DPVO, is at 4.0 cm. So \model's paths have the right shape, and most of the remaining error is length, which depends on the camera. Calibrating on ten labelled examples from the target camera cuts the error by 66\%, to 0.40 m.

\begin{table}[!t]
\centering\scriptsize\setlength{\tabcolsep}{2.5pt}
\resizebox{\columnwidth}{!}{\begin{tabular}{@{}lrrrrrrr@{}}
\toprule
model & params & chal.\ 1w & chal.\ slide & long-tail & DAD & DoTA & DADA \\
 & & $n$=1344 & $n$=1344 & $n$=884 & $n$=314 & $n$=367 & $n$=113 \\
\midrule
\textbf{BADAS-3.0} risk head & 1B & \textbf{0.949} & \textbf{0.892} & \textbf{0.995} & 0.916 & 0.999 & 0.996 \\
\midrule
BADAS-2.0~\cite{badas2} & 300M & 0.940 & 0.876 & 0.994 & \textbf{0.922} & \textbf{0.999} & \textbf{0.996} \\
BADAS-2.0-Flash & 86M & 0.941 & 0.879 & 0.990 & 0.849 & 0.998 & 0.995 \\
BADAS-2.0-Flash-Lite & 22M & 0.933 & 0.867 & 0.984 & 0.870 & 0.998 & 0.992 \\
BADAS-1.0~\cite{badas1} & 300M & 0.925 & 0.861 & 0.960 & 0.885 & 0.998 & 0.989 \\
BADAS-Open & 300M & -- & -- & -- & 0.893 & 0.998 & 0.991 \\
\midrule
Cosmos-Reason2 2B, fine-tuned & 2B & 0.889 & 0.773 & 0.941 & 0.602 & 0.998 & 0.978 \\
Cosmos-Reason2 2B & 2B & 0.694 & 0.667 & 0.755 & 0.141 & 0.983 & 0.939 \\
Gemini 2.5 Pro, fine-tuned & undisclosed & 0.706 & 0.638 & 0.795 & 0.659 & 0.976 & 0.953 \\
Gemini 2.5 Pro & undisclosed & 0.538 & 0.518 & 0.665 & 0.230 & 0.922 & 0.863 \\
Qwen3-VL 2B & 2B & 0.705 & 0.641 & 0.672 & 0.141 & 0.951 & 0.886 \\
\bottomrule
\end{tabular}}
\caption{\textbf{Collision anticipation.} mAP@TTA on Nexar Public, scored under two protocols (single-window, the competition's, and sliding window), clip-level AP elsewhere. Baseline rows are the published predictions of the BADAS-2.0 comparison~\cite{badas2}. DAD has 13 ego-involved positives.}
\label{tab:risk}
\end{table}

\textbf{Collision anticipation.} The risk head is distilled from BADAS-2.0~\cite{badas2}, the previous state of the art. It is ahead of its teacher on Nexar Public under both protocols and on the long-tail set. It ties the teacher on DoTA and DADA-2000 and trails it by 0.006 AP on DAD (Table~\ref{tab:risk}). Its attention falls on the hazard, as BADAS-2.0's does (Figure~\ref{fig:heat}, supplement).\looseness=-1

\subsection{Safety}
\label{sec:safety}
The trajectory benchmarks above score the model against the path the driver took, and in those cases the driver took a safe one. They therefore cannot say whether a planner avoids a collision that is already under way (Section~\ref{sec:related}).

We add two benchmarks built around imminent collisions, on which a planner passes by keeping clear of the hazard. The \emph{Euro NCAP benchmark} replays the protocol's collision-avoidance scenarios in CARLA~\cite{carla}. The \emph{Dashcam Collision Avoidance benchmark} (DCA-Bench) replays real dashcam collisions and near-misses. Each system provides one plan, which is checked against the hazard's scripted or tracked path. No other agent reacts to it.

\emph{Euro NCAP benchmark.} The benchmark is made up of 356 scenario pairs over 15 protocol families, generated from OpenSCENARIO files~\cite{vectorosc} for the Euro NCAP Crash Avoidance Frontal Collisions protocol, v1.1~\cite{euroncapcafc}. Each pair is made up of a scenario from the NCAP protocol and a \emph{hazard-free twin} with the hazard entity removed. 256 pairs are on a straight road, with car, pedestrian, cyclist and motorcycle targets and 100 pairs are junction crossings. Each scenario is rendered using esmini \cite{esmini} and CARLA according to the camera rig required by each system. For each scenario we define the \emph{last-chance instant} as the last timestep where a brake at  6~m/s$^2$ avoids the collision. We evaluate each model at the last-chance instant as well as 0.25 s and 0.5 s before. On the collision variants each system is scored with the Euro NCAP Standard scheme, rescaled from 0 to 1, with the average taken over the three time instants. On the hazard-free twin we report the false positive rate. A false positive is defined as a mean deceleration above 3.5~m/s$^2$ over any 0.5 s period or a lateral movement of more than 2.5 m in either direction. A pair is passed if the collision is avoided and there is no false positive.\looseness=-1

\emph{DCA-Bench.} The Euro NCAP scenarios are schematic. DCA-Bench tests systems on 100 real-world incidents, collisions and near-misses, from the test split of the Nexar Dashcam Collision Prediction Challenge~\cite{nexardataset}. Each clip is annotated with the ego and other road users' trajectories and then cut at the final point before the event when the driver could avoid the collision. Each system sees the two seconds before the cut and produces a trajectory plan. The plan is replayed for 4 s against the reconstructed road users, who do not react, and fails if the ego box comes within 0.3 m of any of them, departs more than 2.5 m sideways from the recorded path or needs more than 8.8 m/s$^2$ of braking from the car's true speed. This stops a plan that assumes a lower speed than the car's true speed from passing by braking harder than a car can. As on Euro NCAP, each incident is paired with a hazard-free twin, cut from the same clip to end before the hazard is visible, and scored with the same false-positive and pair rules. Eighty-nine clips allow such a cut inside the public challenge clip, and the false-positive and pass rates are over these pairs (supplement).\looseness=-1 

\begin{table}[!t]
\centering\scriptsize\setlength{\tabcolsep}{4pt}
\resizebox{\columnwidth}{!}{\begin{tabular}{@{}llrrrrr@{}}
\toprule
 & & \multicolumn{3}{c}{Euro NCAP} & \multicolumn{2}{c}{DCA-Bench} \\
\cmidrule(lr){3-5}\cmidrule(lr){6-7}
model & cameras & score & FP rate (\%) & pass rate (\%) & FP rate (\%) & pass rate (\%) \\
\midrule
Full brake, 6 m/s$^2$ (control) & -- & 1.000 & 100.0 & 0.0 & 93.3 & 6.7 \\
Recorded driver & -- & -- & -- & -- & 0.0 & 44.9 \\
\midrule
\textbf{\model} & 3 front, 150$^\circ$ & 0.365 & 1.4 & \textbf{19.1} & -- & -- \\
\textbf{\model} & 1 front, 75$^\circ$ & 0.296 & 0.0 & 15.4 & 13.5 & \textbf{39.3} \\
Alpamayo~2~Super$^\dagger$ & 6 & 0.202 & 0.1 & 9.8 & 13.1 & 14.2 \\
Alpamayo-R1$^\dagger$ & 4 & 0.116 & 0.0 & 5.8 & 9.7 & 18.2 \\
Alpamayo~1.5$^\dagger$ & 4 & 0.171 & 0.0 & 4.9 & 18.2 & 9.9 \\
SimWAM RL & 1 front, 64$^\circ$ & 0.030 & 31.5 & 0.0 & 92.1 & 0.0 \\
Qwen-Drive SFT & 3 & 0.006 & 0.0 & 0.0 & 24.7 & 7.9 \\
Qwen-Drive RL & 3 & 0.005 & 29.8 & 0.0 & 73.0 & 6.7 \\
SimWAM SFT & 1 front, 64$^\circ$ & 0.001 & 16.3 & 0.0 & 85.4 & 1.1 \\
\bottomrule
\end{tabular}}
\caption{\textbf{Euro NCAP benchmark (356 scenario pairs) and DCA-Bench (100 real incidents).} The Euro NCAP score is the fraction of the protocol's points held, averaged over the three commit instants. FP rate and pass rate (\% of the 356 pairs) are at the last-chance instant. Each planner uses its own camera rig, and \model uses sensor speed and ego history. Rows are sorted by Euro NCAP pass rate, with the best planner per pass-rate column in bold. On DCA-Bench every planner reads the one dashcam, \model with the camera alone, and rates are over the 89 incident--twin pairs. $^\dagger$Mean of 6 runs.}
\label{tab:ncap}
\end{table}

\textbf{Results.} On Euro NCAP (Table~\ref{tab:ncap}) we evaluate \model with sensor speed and ego history in two camera configurations, a 75$^\circ$ view and a 150$^\circ$ view stitched from three cameras. The wider view can capture hazards coming from the side.\looseness=-1

\model leads every planner we tested with either configuration, although it was never trained on simulated data. With the 75$^\circ$ view its score is 0.296, against 0.202 for the next best planner. With the 150$^\circ$ view, which is outside its training data, its score is 0.365 and its pass rate rises from 15.4\% to 19.1\%. Qwen-Drive and SimWAM score at most 0.030.\looseness=-1

On DCA-Bench (last two columns), every planner reads the one dashcam of the public challenge clips. The other planners also get ego history, and \model reads the camera alone. \model passes 39\% of the pairs, over twice any other planner (0 to 18\%). The recorded driver passes 45\%, because some incidents were near-misses. Seven of \model's 12 false positives come from planning at a speed below the true one, since it has no speed input. A control that always brakes at 6~m/s$^2$ passes 98 of the 100 incidents but fails 93\% of the twins.\looseness=-1

\begin{table}[!t]
\centering\scriptsize\setlength{\tabcolsep}{4pt}
\resizebox{\columnwidth}{!}{\begin{tabular}{@{}llrrrrr@{}}
\toprule
 & & \multicolumn{3}{c}{Euro NCAP} & \multicolumn{2}{c}{DCA-Bench} \\
\cmidrule(lr){3-5}\cmidrule(lr){6-7}
\model & cameras & score & FP rate (\%) & pass rate (\%) & FP rate (\%) & pass rate (\%) \\
\midrule
-- & 1 front, 75$^\circ$ & 0.193 & 0.0 & 4.8 & 13.5 & 20.2 \\
with selector fine-tune & 1 front, 75$^\circ$ & 0.296 & 0.0 & 15.4 & 13.5 & 39.3 \\
\midrule
-- & 3 front, 150$^\circ$ & 0.224 & 0.0 & 6.7 & -- & -- \\
with selector fine-tune & 3 front, 150$^\circ$ & 0.365 & 1.4 & 19.1 & -- & -- \\
\bottomrule
\end{tabular}}
\caption{\textbf{Effect of the safety selector.} The same \model checkpoint on both benchmarks, scored on the candidate the trajectory head ranks first (--) and on the candidate the safety selector chooses.}
\label{tab:selector}
\end{table}

\textbf{Effect of the safety selector.} Table~\ref{tab:selector} isolates the safety-aware training of Section~\ref{sec:training}. The backbone and every head are frozen when the safety selector is trained, so both rows have the same six candidates and differ only in which one is chosen. Ranked by the trajectory head alone, which is trained to imitate the driver, \model passes 20\% of the DCA-Bench pairs, just above the best other planner (18\%). The safety selector nearly doubles this to 39\%, with the false-positive rate on the twins unchanged (13\%). On Euro NCAP, it raises the pass rate 3.2$\times$ with the 75$^\circ$ view and 2.9$\times$ with the 150$^\circ$ view.\looseness=-1

\subsection{NavSim}
\label{sec:navsim}

We evaluate \model on the planning benchmark NavSim v1~\cite{navsim2024}, scored by the PDM score (PDMS). First we keep \model frozen and fit six calibration parameters against ADE on 4,000 navtrain scenes. We then train a NavSim selector on navtrain, a regression model that predicts each candidate's PDMS and picks the highest. It is separate from the safety selector and used for this benchmark alone. Second, we fine-tune \model on navtrain with LoRA~\cite{lora}, then apply GRPO to the full model and the NavSim selector. The comparison planners are all trained on navtrain, some of them with RL. We also report the best-of-6 PDMS, the score of the best candidate per scene.\looseness=-1 

\textbf{Results.} With the frozen model and six calibration parameters, \model reaches 89.0 PDMS from one camera (Table~\ref{tab:navsim}, supplement). Among the non-RL planners we compare against, only SimWAM SFT is higher, at 90.3. LoRA fine-tuning lifts \model to 90.0 and GRPO to 91.1. SimWAM scores 91.9 after RL, and the other RL planners score 89.1 to 91.0. The best-of-6 PDMS of \model's candidates is 96.3 after GRPO (Table~\ref{tab:navsimbestofn}, supplement), so the selection limits the score more than the candidate set does.\looseness=-1

\section{Conclusion and limitations}
\label{sec:conclusion}
\label{sec:limitations}
From one forward camera, the 1B-parameter \model returns speed, metric ego-pose, collision risk and six candidate futures in under 50 ms. On trajectory forecasting it is on par with Alpamayo~2~Super on the camera they share and ahead on the other four sets. Its speed and ego-pose heads have the lowest mean error among the systems we compare against. On two new safety benchmarks it has a higher pass rate than any other planner we tested.\looseness=-1

\textbf{Limitations.} The short, single-camera input that makes \model fast also limits it. The model sees two seconds from a forward camera with no state between windows and is not trained in closed loop. So it cannot anticipate hazards beside it, behind it or out of view, and drift accumulates over long drives. Both safety benchmarks are open-loop, and other road users do not react to the plan. The backbone and the safety selector are trained on proprietary dashcam data, so their training cannot be reproduced, although we release the weights. DAD has only 13 ego-involved positives, so its AP is noisy.\looseness=-1

\section*{LLM Usage Statement}
We used Claude (Anthropic), through the Claude Code tool, as a writing and coding assistant. For the text, it drafted and edited prose, \LaTeX{} and captions under the authors' direction. For figures and tables, it wrote the plotting and table-generation scripts. For experiments, it wrote and changed evaluation code, prepared job configurations and helped analyse results. The authors chose the research questions, experiments and claims, and checked all results, numbers and citations in the paper.

{
    \small
    \bibliographystyle{ieeenat_fullname}
    \bibliography{main,navsim}
}
\clearpage
\appendix
\section{Supplementary Material}
\label{sec:supp}
\renewcommand{\topfraction}{0.9}\renewcommand{\bottomfraction}{0.9}\renewcommand{\textfraction}{0.1}\renewcommand{\floatpagefraction}{0.8}
\renewcommand{\dbltopfraction}{0.9}\renewcommand{\dblfloatpagefraction}{0.8}\setcounter{totalnumber}{6}\setcounter{dbltopnumber}{4}

\subsection{Ego-pose after similarity alignment}
\begin{figure*}[!t]
  \centering
  \includegraphics[width=\textwidth]{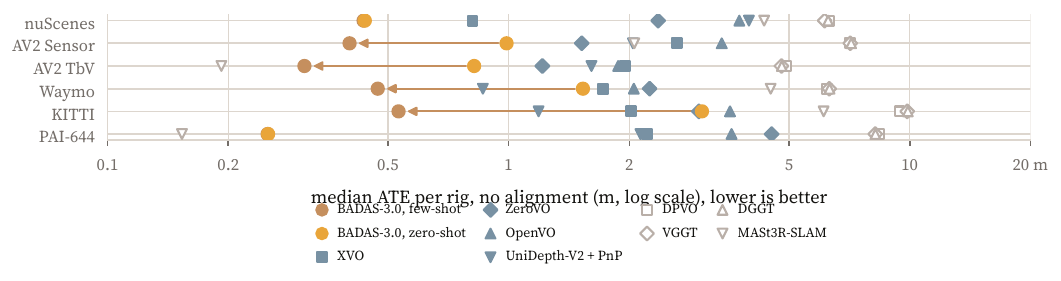}
  \caption{\textbf{Median ATE per rig, no alignment}, log axis. Gold zero-shot, sandstone after ten calibration windows; blue metric VO; hollow scale-free methods scored as delivered. Arrows mark what the per-camera scalar buys.}
  \label{fig:egorigs}
\end{figure*}

\begin{table*}[!t]
\centering\scriptsize\setlength{\tabcolsep}{3.5pt}
\resizebox{\textwidth}{!}{\begin{tabular}{@{}llrrrrrrrr@{}}
\toprule
method & scale & nuSc & AV2-S & AV2-TbV & Waymo & KITTI & PAI-644 & mean$_6$ & runtime \\
\midrule
\textbf{BADAS-3.0}, speed head, zero-shot & metric & 0.033 & 0.031 & \textbf{0.023} & 0.040 & 0.057 & \textbf{0.024} & \textbf{0.035} & 43 ms \\
\textbf{BADAS-3.0}, speed head, few-shot & metric & 0.034 & 0.032 & \textbf{0.023} & 0.041 & 0.057 & 0.025$^{\ddagger}$ & 0.035$^{\ddagger}$ & 43 ms \\
\midrule
XVO~\cite{xvo} & metric & 0.063 & 0.102 & 0.108 & 0.063 & 0.118 & 0.253 & 0.118 & 190 ms \\
ZeroVO~\cite{zerovo} & metric & 0.074 & 0.132 & 0.217 & 0.056 & 0.118 & 0.479 & 0.179 & 9.4 s \\
OpenVO~\cite{openvo} & metric & 0.060 & 0.091 & 0.123 & 0.038 & 0.076 & 0.478 & 0.145 & 5.8 s \\
UniDepth-V2 + PnP~\cite{unidepth} & metric & 1.685 & 0.258 & 0.392 & 0.256 & 0.508 & 0.397 & 0.583 & 1.0 s \\
\midrule
DPVO~\cite{dpvo} & scale-free & \textbf{0.028} & \textbf{0.012} & 0.024 & \textbf{0.023} & \textbf{0.034} & 0.119 & 0.040 & 667 ms \\
VGGT~\cite{vggt} & scale-free & 0.047 & 0.039 & 0.031 & 0.033 & 0.074 & 0.176 & 0.067 & 786 ms \\
DGGT~\cite{dggt} & scale-free & 0.048 & 0.033 & 0.028 & 0.041 & 0.084 & 0.102 & 0.056 & 1.1 s \\
MASt3R-SLAM~\cite{mast3rslam} & scale-free & 0.699 & 0.110 & 0.066 & 0.239 & 0.232 & 0.076 & 0.237 & 10.9 s \\
\bottomrule
\end{tabular}}
\caption{\textbf{Two-second ego-pose, median ATE in metres after a per-window similarity fit} of rotation, translation and scale. Same windows, rows and runtime column as Table~\ref{tab:ego}. Mean$_6$ is the mean over the six columns. $^{\ddagger}$\,few-shot cell fitted with the earlier rule (Section~\ref{sec:calibration}).}
\label{tab:egosim3}
\end{table*}
\begin{figure}[!htbp]
  \centering
  \includegraphics[width=\columnwidth]{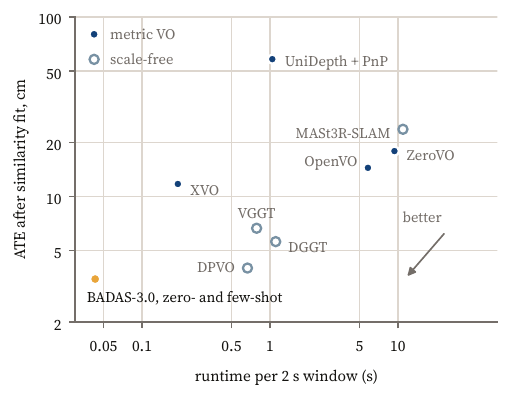}
  \caption{\textbf{Shape error against runtime.} Mean$_6$ of the per-rig median ATE after a per-window similarity fit, in centimetres. Both \model rows collapse onto one point at 3.5 cm.}
  \label{fig:egocostsim3}
\end{figure}
Table~\ref{tab:egosim3} and Figure~\ref{fig:egocostsim3} rescore the windows of Table~\ref{tab:ego} after a per-window similarity fit, which tests the approximation of Section~\ref{sec:heads}. Under alignment both \model rows produce nearly identical shape error, 3.5 cm, so most of the camera-only gap in the unaligned table is one number per rig. The fit also absorbs heading and origin error, so this bounds the scale effect. In this shape column the geometry systems reach 4.0 to 6.7 cm on the six-rig mean; DPVO is finer on the survey-grade rigs and coarser on the fisheye PhysicalAI rig. Figure~\ref{fig:egorigs} shows the unaligned per-rig values as bars.

\begin{figure}[!htbp]
  \centering
  \includegraphics[width=\columnwidth]{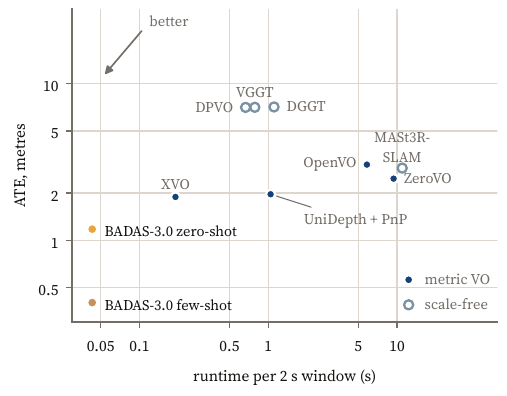}
  \caption{\textbf{Two-second ego-pose: metres against seconds.} Mean$_6$ of the per-rig median ATE with no alignment, against per-window runtime (Table~\ref{tab:ego}). \model zero-shot and few-shot, against metric visual odometry (filled) and scale-free reconstruction (hollow) scored as delivered.}
  \label{fig:egocost}
\end{figure}

\subsection{Speed per rig}
Table~\ref{tab:speed} gives the speed error per rig, with every row scored on the same windows.
\begin{table}[!htbp]
\centering\scriptsize\setlength{\tabcolsep}{3pt}
\resizebox{\columnwidth}{!}{\begin{tabular}{@{}lrrrrrrr@{}}
\toprule
method & nuSc & AV2-S & AV2-TbV & Waymo & KITTI & PAI-100c & mean$_6$ \\
\midrule
\textbf{BADAS-3.0} speed head, zero-shot & \textbf{0.50} & 0.84 & 0.75 & 1.68 & 3.22 & 0.55 & 1.26 \\
\textbf{BADAS-3.0} speed head, few-shot & \textbf{0.50} & \textbf{0.46} & \textbf{0.41} & \textbf{0.70} & \textbf{0.56} & \textbf{0.52} & \textbf{0.52} \\
\midrule
XVO & 1.02 & 2.14 & 1.95 & 2.31 & 2.32 & 4.36 & 2.35 \\
ZeroVO & 1.94 & 2.76 & 2.72 & 2.53 & 3.27 & 6.67 & 3.31 \\
OpenVO & 3.52 & 7.05 & 4.87 & 3.51 & 3.64 & 4.99 & 4.60 \\
UniDepth-V2 + PnP & 17.09 & 5.93 & 4.39 & 65.90 & 3.82 & 7.15 & 17.38 \\
\bottomrule
\end{tabular}}
\caption{\textbf{Speed at $t_0$ from one camera, mean absolute error in m/s.} Baselines' speed is the magnitude of their own last-step translation. Every row is scored on the same windows, each rig's first ten (the few-shot calibration windows) excluded.}
\label{tab:speed}
\end{table}

\subsection{Full drives}
\label{sec:fulldrives}
\begin{table*}[!t]
\centering\scriptsize\setlength{\tabcolsep}{3.5pt}
\resizebox{\textwidth}{!}{\begin{tabular}{@{}lrrrrrrrr@{}}
\toprule
 & \multicolumn{4}{c}{no alignment} & \multicolumn{4}{c}{similarity fit} \\
\cmidrule(lr){2-5}\cmidrule(l){6-9}
method & PAI-644 & AV2 & Waymo & Prop. & PAI-644 & AV2 & Waymo & Prop. \\
\midrule
\textbf{BADAS-3.0}, stitched, zero-shot & 1.82 & 2.44 & 10.07 & \textbf{4.94} & \textbf{0.33} & 0.54 & 0.52 & \textbf{0.91} \\
\textbf{BADAS-3.0}, stitched, few-shot & \textbf{1.62}$^{\ddagger}$ & \textbf{2.37} & 4.70 & 10.65$^{\ddagger}$ & 0.34$^{\ddagger}$ & 0.54 & 0.68 & 0.94$^{\ddagger}$ \\
\textbf{BADAS-3.0}, chained, zero-shot & 2.11 & 2.82 & 12.10 & 6.03 & 0.42 & 0.66 & 0.76 & 1.15 \\
\textbf{BADAS-3.0}, chained, few-shot & 1.96$^{\ddagger}$ & 2.73 & \textbf{4.15} & 11.80$^{\ddagger}$ & 0.44$^{\ddagger}$ & 0.64 & 0.91 & 1.12$^{\ddagger}$ \\
\textbf{BADAS-3.0}, stitched, speed sensor (ref.) & 1.03 & 1.27 & 2.53 & 1.98 & 0.22 & 0.36 & 0.44 & 0.66 \\
\midrule
XVO & 15.18 & 20.79 & 13.07 & 28.54 & 2.30 & 1.45 & 0.80 & 3.62 \\
ZeroVO & -- & 9.11 & 17.58 & 46.80 & -- & 1.43 & 0.68 & 5.01 \\
OpenVO & 20.42 & 16.05 & 15.27 & 39.26 & 4.55 & 0.90 & 0.45 & 6.02 \\
UniDepth-V2 + PnP & 13.84 & 9.96 & 4.42 & 53.17 & 2.09 & 1.33 & 0.95 & 9.38 \\
\midrule
DPVO (scale-free) & 49.59 & 39.74 & 45.62 & 80.43 & 1.77 & \textbf{0.18} & 0.47 & 0.93 \\
VGGT (scale-free) & 48.83 & 39.95 & 46.97 & 80.93 & 0.76 & 0.36 & \textbf{0.38} & 1.31 \\
DGGT (scale-free) & -- & -- & -- & 81.02 & -- & -- & -- & 1.35 \\
\bottomrule
\end{tabular}}
\caption{\textbf{Full drives, median ATE in metres over 15-second segments}, with no alignment and after one similarity fit per segment. Chained joins consecutive two-second windows by a rigid fit on their overlap; stitched uses the learned stitcher. Few-shot rows fit one speed factor per set (Equation~\ref{eq:gate}) on its first ten usable segments, which are not scored; $^{\ddagger}$\,fitted with the earlier rule (Section~\ref{sec:calibration}). The speed-sensor row is a reference and is not ranked. --: no result.}
\label{tab:full}
\end{table*}
\textbf{Task.} Each method reads a 15-second front-camera segment and returns the ego path over the whole segment, expressed in the frame of its final pose. ATE is computed over the full path, and we report the median over segments with no alignment and after a similarity fit, as in Table~\ref{tab:ego}. The sets are PAI-644 (one segment per clip), 148 Argoverse~2 Sensor segments, 100 Waymo segments and 68 proprietary dashcam segments, the last with ground truth from 3D reconstruction. The baselines read the whole segment with their own sequence pipelines.

\textbf{Chaining and stitching.} \model sees two seconds at a time and keeps no state, so a long path has to be assembled from windows. The simplest assembly, \emph{chained}, slides the 16-frame window one frame at a time and joins consecutive windows by a rigid fit on their 1.875 s overlap. The \emph{stitcher} replaces this with a small learned model. Windows are taken every 0.5 s, and for each one the stitcher reads the 128 pooled backbone tokens, the ego-pose head's acceleration and curvature, the speed head's $v_0$ and the time since the previous window. A causal transformer over the last eight windows (three layers, width 256, 10.7M parameters) outputs a yaw-rate correction and a log standard deviation for each of the newest window's 19 steps. The yaw rates of all overlapping windows are fused by inverse variance on a 0.1 s grid and integrated once with their mean speed, with the yaw rate set to zero below 0.5 m/s. The backbone and heads are frozen, so the two-second outputs are the same with or without the stitcher.

\textbf{Training.} The stitcher is trained on 37.6k PhysicalAI-AV and 2.6k proprietary dashcam long clips with geometric camera-warp augmentation. The loss is a Huber loss on the wrapped cumulative yaw at 0.5, 1, 2, 4 and 8 s, weighted five times on turns, plus smaller position, uncertainty and climb-rate terms. It trains for six epochs on one A100.

\textbf{Results.} Zero-shot, the stitcher lowers the error of chaining on every set in both columns of Table~\ref{tab:full}, for example from 2.11 to 1.82 m on PAI-644 and from 2.82 to 2.44 m on AV2. On PAI-644, AV2 and the proprietary set zero-shot \model has 3.7 to 7.6$\times$ lower unaligned error than the best metric pipeline. On Waymo the zero-shot speed scale is off, as in Table~\ref{tab:ego}, and over 15 s the error grows to 10.1 m; ten calibration segments bring it to 4.7 m stitched and 4.2 m chained, the chained row below UniDepth-V2 + PnP at 4.4 m. On AV2 the zero-shot scale is already right: the gate keeps it for the stitched row (2.37 m on the scored segments) and the chained factor is 0.93 (2.73 m against 2.81 m zero-shot on the same segments). On the proprietary set the few-shot rows, fitted with the earlier rule, raise the error, from 4.9 to 10.7 m stitched. With sensor speed the stitched path is within 1.0 to 2.5 m, so most of the remaining error is speed. After the similarity fit, scale-free reconstruction has the lowest shape error on AV2 and Waymo, and the stitcher on PAI-644 and the proprietary set.

\subsection{Where the risk head looks}
\begin{figure}[!htbp]
  \centering
  \includegraphics[width=\columnwidth]{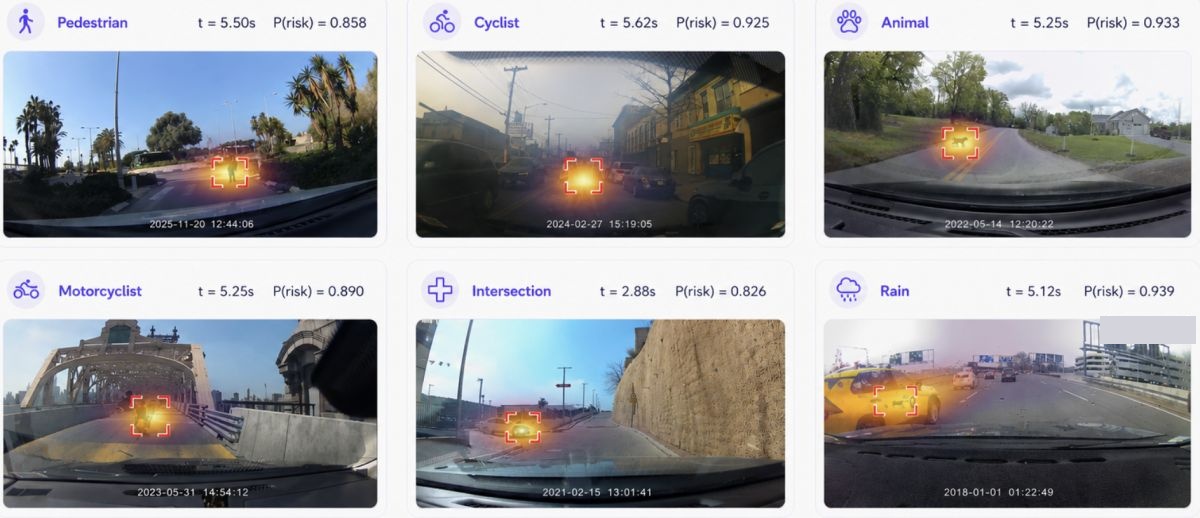}
  \caption{\textbf{One clip from each of six of the ten long-tail groups, at the moment of the alert}, collision probability 0.83 to 0.94. The attribution peak sits on the hazard; brackets mark it.}
  \label{fig:heat}
\end{figure}

Figure~\ref{fig:heat} shows where the risk head's attribution falls on clips from the long-tail set at the moment of the alert.

\subsection{Anchors, deltas and the closest-mode loss}
\begin{figure*}[!t]
  \centering
  \includegraphics[width=0.8\textwidth]{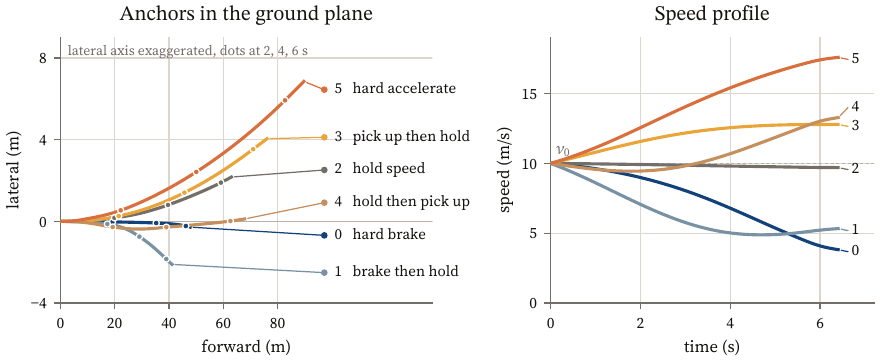}
  \caption{\textbf{The six mode anchors, integrated.} Each anchor's control sequence rolled out through the unicycle model from $v_0=10$ m/s. Left: the paths in the ground plane, with dots at 2, 4 and 6 s; the lateral axis is exaggerated. Right: the speed profile each anchor implies, from hard braking (0) to hard acceleration (5).}
  \label{fig:anchors_xy}
\end{figure*}
\begin{figure*}[!t]
  \centering
  \includegraphics[width=0.9\textwidth]{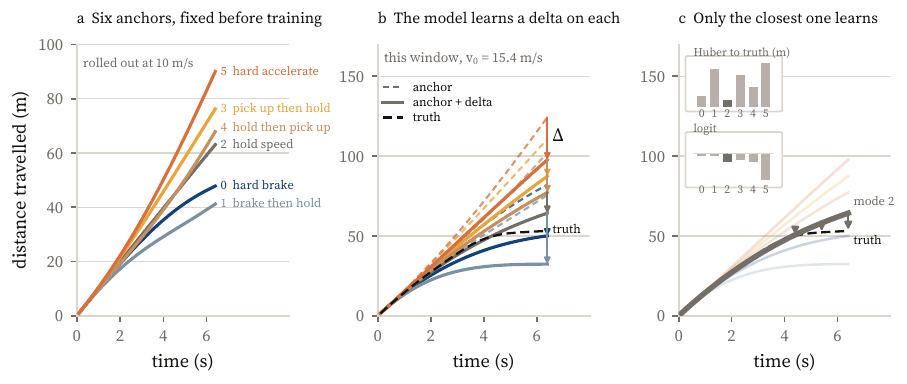}
  \caption{\textbf{Six anchors, six deltas, one truth.} (a) Six fixed control sequences, chosen once by $k$-means over the training set and never updated, rolled out at 10 m/s: they span hard brake to hard acceleration. (b) For a real window the model adds a learned delta to each anchor (dashed to solid); truth dashed orange. (c) Only the mode closest to the truth is trained, its path pulled onto the truth and its score pushed up; the other five receive no path gradient, so braking scenes teach the braking anchors and cruising scenes the cruising one, and the six stay diverse.}
  \label{fig:anchors}
\end{figure*}
Figure~\ref{fig:anchors_xy} shows the six anchors as paths and speed profiles, and Figure~\ref{fig:anchors} shows them, the deltas the head adds, and the single closest-mode loss that keeps them apart.

\subsection{Trajectory means with uncertainty}
\begin{table*}[!t]
\centering\scriptsize\setlength{\tabcolsep}{3.5pt}
\resizebox{\textwidth}{!}{\begin{tabular}{@{}lrrrrrr@{}}
\toprule
model & nuSc & AV2-S & AV2-TbV & Waymo & PAI-644 & mean$_5$ \\
\midrule
\textbf{BADAS-3.0}, speed head, zero-shot & 1.22$\pm$0.12 & 1.44$\pm$0.14 & 1.29$\pm$0.10 & 3.19$\pm$0.40 & 1.27$\pm$0.14 & 1.682 \\
\textbf{BADAS-3.0}, speed head, few-shot & 1.22$\pm$0.12 & 1.23$\pm$0.15 & 1.06$\pm$0.09 & 1.52$\pm$0.20 & 1.27$\pm$0.14 & 1.261 \\
\textbf{BADAS-3.0}, speed sensor & 0.90$\pm$0.09 & 1.01$\pm$0.13 & 0.94$\pm$0.08 & 1.00$\pm$0.11 & 0.93$\pm$0.06 & 0.955 \\
\textbf{BADAS-3.0}, speed sensor + ego history & 0.89$\pm$0.09 & 1.00$\pm$0.13 & 0.95$\pm$0.09 & 0.99$\pm$0.11 & 0.93$\pm$0.06 & 0.952 \\
Alpamayo 2 Super & 2.35$\pm$0.30 & 1.98$\pm$0.34 & 1.77$\pm$0.20 & 1.41$\pm$0.24 & 0.96$\pm$0.09 & 1.694 \\
Alpamayo-R1 & 2.14$\pm$0.22 & 2.16$\pm$0.41 & 2.08$\pm$0.22 & 1.69$\pm$0.23 & 1.03$\pm$0.09 & 1.819 \\
Alpamayo 1.5 & 2.12$\pm$0.22 & 2.29$\pm$0.40 & 1.89$\pm$0.19 & 1.44$\pm$0.24 & 0.95$\pm$0.10 & 1.738 \\
Alpamayo 2 Edge & 1.74$\pm$0.19 & 1.74$\pm$0.26 & 1.97$\pm$0.16 & 1.26$\pm$0.16 & 1.27$\pm$0.12$^{\dagger}$ & 1.597$^{\dagger}$ \\
\bottomrule
\end{tabular}}
\caption{\textbf{minADE$_6$ over a 6.4-second horizon with a 95\% half-width per column} ($1.96\,\sigma/\sqrt{n}$ over windows), and mean$_5$ over the five sets. $^\dagger$As in Table~\ref{tab:traj}.}
\label{tab:trajci}
\end{table*}
Table~\ref{tab:trajci} gives the per-column half-widths behind Table~\ref{tab:traj}, together with the five-set mean.

\subsection{Baseline inputs and checks}
\label{sec:baselineinputs}
\begin{table*}[!t]
\centering\scriptsize\setlength{\tabcolsep}{3.5pt}
\resizebox{\textwidth}{!}{\begin{tabular}{@{}llllll@{}}
\toprule
slot (index) & rig heading & nuScenes & AV2 Sensor and TbV & Waymo & PAI-644 \\
\midrule
cross\_left (0) & 67$^\circ$ & CAM\_FRONT\_LEFT (55$^\circ$) & ring\_front\_left (45$^\circ$) & FRONT\_LEFT (45$^\circ$) & cross\_left \\
front\_wide (1) & 0$^\circ$ & CAM\_FRONT & ring\_front\_center & FRONT & front\_wide \\
cross\_right (2) & $-67^\circ$ & CAM\_FRONT\_RIGHT ($-58^\circ$) & ring\_front\_right ($-45^\circ$) & FRONT\_RIGHT ($-45^\circ$) & cross\_right \\
rear\_left (3) & 152$^\circ$ & CAM\_BACK\_LEFT (109$^\circ$) & ring\_rear\_left (153$^\circ$) & SIDE\_LEFT (90$^\circ$) & rear\_left \\
rear\_right (5) & $-152^\circ$ & CAM\_BACK\_RIGHT ($-113^\circ$) & ring\_rear\_right ($-153^\circ$) & SIDE\_RIGHT ($-90^\circ$) & rear\_right \\
front\_tele (6) & 0$^\circ$, 30$^\circ$ FOV & central crop of CAM\_FRONT & central crop of ring\_front\_center & central 30$^\circ$ crop of FRONT & front\_tele \\
\bottomrule
\end{tabular}}
\caption{\textbf{Camera per Alpamayo slot.} Optical-axis heading in the ego frame from each dataset's calibration (positive to the left). Alpamayo~2~Super reads all six slots; Alpamayo-R1, 1.5 and 2~Edge read slots 0, 1, 2 and 6. Waymo has no rear-facing camera, so its side cameras fill the rear slots. Every view is centre-cropped to 1.8:1 and resized to $320\times576$; four frames per camera at $t_0-0.3$, $-0.2$, $-0.1$ and $0$ s, the nearest native frame per camera.}
\label{tab:inputs}
\end{table*}

\begin{table}[!htbp]
\centering\scriptsize\setlength{\tabcolsep}{3.5pt}
\resizebox{\columnwidth}{!}{\begin{tabular}{@{}lrrrrr@{}}
\toprule
model & nuSc & AV2-S & AV2-TbV & Waymo & PAI-644 \\
\midrule
constant velocity & 3.714 & 4.094 & 3.800 & 3.497 & 6.674 \\
\midrule
\textbf{BADAS-3.0}, head's own pick & \textbf{2.610} & \textbf{2.794} & \textbf{2.714} & \textbf{2.713} & \textbf{2.726} \\
Alpamayo 2 Super & 5.036 & 4.498 & 3.923 & \textbf{2.723} & \textbf{2.288} \\
Alpamayo-R1 & 4.777 & 4.922 & 4.152 & \textbf{3.108} & \textbf{2.301} \\
Alpamayo 1.5 & 4.395 & 4.374 & 3.867 & \textbf{2.543} & \textbf{2.075} \\
\bottomrule
\end{tabular}}
\caption{\textbf{One trajectory against constant velocity}, ADE over 6.4 s in metres on the windows of Table~\ref{tab:traj}: \model's head's own pick (Table~\ref{tab:traj}'s top-1), Alpamayo's first sample. Constant velocity extrapolates the last 0.1 s of ego history. Bold: lower error than constant velocity. Averaged over its six samples instead (the top-1 of Table~\ref{tab:traj}), Alpamayo~2~Super is ahead of constant velocity on Argoverse~2 Sensor (3.97 against 4.09 m) and Argoverse~2 TbV (3.70 against 3.80 m), and Alpamayo~1.5 on Argoverse~2 TbV by 5 mm; Alpamayo~2~Super, Alpamayo-R1 and Alpamayo~1.5 trail it on nuScenes, Alpamayo-R1 and Alpamayo~1.5 on Argoverse~2 Sensor, and Alpamayo-R1 on Argoverse~2 TbV.}
\label{tab:cv}
\end{table}

\begin{table}[!htbp]
\centering\scriptsize\setlength{\tabcolsep}{3.5pt}
\resizebox{\columnwidth}{!}{\begin{tabular}{@{}llrrr@{}}
\toprule
set & run & minADE$_6$ & first sample & mean of six \\
\midrule
nuSc & Alpamayo 1.5, seed 42 & 2.117 & 4.395 & 4.292 \\
nuSc & Alpamayo 1.5, seed 43 & 1.859 & 4.677 & 4.458 \\
nuSc & Alpamayo 1.5, seed 44 & 1.924 & 3.687 & 4.097 \\
nuSc & Alpamayo 1.5, three seeds & 1.967$\pm$0.134 & 4.253$\pm$0.510 & 4.282$\pm$0.181 \\
\midrule
PAI-644 & Alpamayo 1.5, seed 42 & 0.954 & 2.075 & 1.987 \\
PAI-644 & Alpamayo 1.5, seed 43 & 0.921 & 2.037 & 1.991 \\
PAI-644 & Alpamayo 1.5, seed 44 & 0.900 & 2.021 & 1.993 \\
PAI-644 & Alpamayo 1.5, three seeds & 0.925$\pm$0.027 & 2.044$\pm$0.028 & 1.990$\pm$0.003 \\
PAI-644 & Alpamayo 2 Super, six cameras & 0.957 & 2.288 & 2.127 \\
PAI-644 & Alpamayo 2 Super, four cameras & 0.958 & 2.311 & 2.145 \\
\bottomrule
\end{tabular}}
\caption{\textbf{Sampling seed and camera count}, metres at 6.4 s on one machine. Table~\ref{tab:traj} uses seed 42 and six cameras for Alpamayo~2~Super; the three-seed row gives mean and standard deviation.}
\label{tab:seeds}
\end{table}
Every view an Alpamayo model reads comes from the nearest physical camera of the rig (Table~\ref{tab:inputs}). AV2's rear cameras point within 1$^\circ$ of the PhysicalAI-AV rig's, nuScenes' about 40$^\circ$ further forward. The audit confirms that each adapter passes the 16-pose ego history the models specify, expressed relative to the pose at $t_0$, that each frame is the nearest native frame to its target time, and that predicted paths carry no rotation, flip or scale error against the ground truth. A single trajectory separates the rigs (Table~\ref{tab:cv}): on Waymo and PAI-644 every model beats constant velocity, while on nuScenes and both Argoverse sets only \model does. Sampling spread depends on the rig as well (Table~\ref{tab:seeds}): across three seeds Alpamayo~1.5 moves by 0.03 m on PAI-644 and by up to 1 m in the ADE of its first sample on nuScenes, and Alpamayo~2~Super scores the same with six cameras as with four.

\subsection{Trajectory at 5 s, with Qwen-Drive}
\begin{table*}[!t]
\centering\scriptsize\setlength{\tabcolsep}{3.5pt}
\resizebox{\textwidth}{!}{\begin{tabular}{@{}llrrrrrrr@{}}
\toprule
model & input & params & nuSc & AV2-S & AV2-TbV & Waymo & PAI-644 & mean$_5$ \\
\midrule
\textbf{BADAS-3.0}, speed head, zero-shot & 1 cam & 1B & 0.928 & 1.114 & 1.023 & 2.690 & 0.983 & 1.347 \\
\textbf{BADAS-3.0}, speed head, few-shot & 1 cam & 1B & 0.916 & 0.939 & 0.798 & 1.221 & 0.983 & 0.971 \\
\textbf{BADAS-3.0}, speed sensor & 1 cam & 1B & \textbf{0.610} & \textbf{0.653} & \textbf{0.637} & 0.672 & 0.637 & 0.642 \\
\textbf{BADAS-3.0}, speed sensor + ego history & 1 cam & 1B & 0.612 & 0.654 & 0.643 & \textbf{0.661} & 0.631 & \textbf{0.640} \\
\midrule
Alpamayo 2 Super~\cite{alpamayocards} & 6 cam & 34B & 1.582 & 1.305 & 1.179 & 0.923 & 0.623 & 1.122 \\
Alpamayo-R1~\cite{alpamayor1} & 4 cam & 10B & 1.509 & 1.455 & 1.420 & 1.131 & 0.671 & 1.237 \\
Alpamayo 1.5~\cite{alpamayocards} & 4 cam & 10.5B & 1.410 & 1.529 & 1.302 & 0.955 & \textbf{0.603} & 1.160 \\
Alpamayo 2 Edge~\cite{alpamayoedge} & 4 cam & 2B & 1.124 & 1.102 & 1.206 & 0.792 & 0.772$^{\dagger}$ & 0.999$^{\dagger}$ \\
\midrule
Qwen-Drive 4B, SFT~\cite{qwendrive} & 3 cam & 4B & 1.950 & 1.997 & 1.848 & 1.799 & 1.861 & 1.891 \\
Qwen-Drive 4B, RL~\cite{qwendrive} & 3 cam & 4B & 1.986 & 2.104 & 1.873 & 1.892 & 1.765 & 1.924 \\
\bottomrule
\end{tabular}}
\caption{\textbf{minADE$_6$ at 5.0 s, metres.} Same windows and conventions as Table~\ref{tab:traj}; Qwen-Drive predicts 5 s, reads three cameras and receives the constant no-route command. $^\dagger$As in Table~\ref{tab:traj}: NVIDIA's anchor windows of the 644 clips. SimWAM is omitted because its released adapter does not anchor predictions at the ego origin.}
\label{tab:traj5}
\end{table*}
Table~\ref{tab:traj5} scores the same windows at a 5 s horizon, which lets Qwen-Drive, whose released head stops at 5 s, join the comparison.

\subsection{Euro NCAP benchmark at each commit instant}
Table~\ref{tab:ncapfull} expands Table~\ref{tab:ncap} to every commit instant: the protocol score, the pairs passed and the false-positive rate.
\begin{table*}[!t]
\centering\scriptsize\setlength{\tabcolsep}{3.5pt}
\resizebox{\textwidth}{!}{\begin{tabular}{@{}llrrrrrrrrrr@{}}
\toprule
model & cameras & \multicolumn{4}{c}{score} & \multicolumn{3}{c}{pairs passed} & \multicolumn{3}{c}{FP rate (\%)} \\
\cmidrule(lr){3-6}\cmidrule(lr){7-9}\cmidrule(lr){10-12}
 & & $-0.5$ s & $-0.25$ s & $0$ s & mean & $-0.5$ s & $-0.25$ s & $0$ s & $-0.5$ s & $-0.25$ s & $0$ s \\
\midrule
Full brake, 6 m/s$^2$ (control) & -- & 1.000 & 1.000 & 1.000 & 1.000 & 0 & 0 & 0 & 100.0 & 100.0 & 100.0 \\
\midrule
\textbf{\model} & 3 front, 150$^\circ$ & 0.401 & 0.423 & 0.272 & 0.365 & 143 & 131 & 68 & 2.5 & 2.5 & 1.4 \\
\textbf{\model} & 1 front, 75$^\circ$ & 0.357 & 0.317 & 0.214 & 0.296 & 135 & 115 & 55 & 0.0 & 0.0 & 0.0 \\
\model, head's own pick & 3 front, 150$^\circ$ & 0.296 & 0.229 & 0.146 & 0.224 & 111 & 72 & 24 & 0.3 & 0.0 & 0.0 \\
Alpamayo~2~Super$^\dagger$ & 6 & 0.220 & 0.222 & 0.164 & 0.202 & 85.7 & 71.7 & 35.0 & 0.5 & 0.2 & 0.1 \\
\model, head's own pick & 1 front, 75$^\circ$ & 0.246 & 0.210 & 0.122 & 0.193 & 90 & 61 & 17 & 0.0 & 0.0 & 0.0 \\
Alpamayo~1.5$^\dagger$ & 4 & 0.227 & 0.186 & 0.101 & 0.171 & 89.2 & 61.3 & 17.5 & 0.7 & 0.4 & 0.0 \\
Alpamayo-R1$^\dagger$ & 4 & 0.143 & 0.121 & 0.084 & 0.116 & 59.5 & 42.8 & 20.7 & 0.0 & 0.0 & 0.0 \\
SimWAM RL & 1 front, 64$^\circ$ & 0.048 & 0.034 & 0.009 & 0.030 & 2 & 1 & 0 & 87.6 & 59.8 & 31.5 \\
Qwen-Drive SFT & 3 & 0.006 & 0.006 & 0.006 & 0.006 & 0 & 0 & 0 & 0.0 & 0.0 & 0.0 \\
Qwen-Drive RL & 3 & 0.007 & 0.005 & 0.003 & 0.005 & 0 & 0 & 0 & 51.1 & 43.0 & 29.8 \\
SimWAM SFT & 1 front, 64$^\circ$ & 0.002 & 0.001 & 0.000 & 0.001 & 1 & 1 & 0 & 75.0 & 41.6 & 16.3 \\
\bottomrule
\end{tabular}}
\caption{\textbf{Euro NCAP benchmark, every column.} The protocol score, pairs passed (of 356) and false-positive rate at 0.5 s and 0.25 s before the last-chance instant and at the instant itself; this table is ranked by the mean of the three scores. $^\dagger$Mean of 6 runs.}
\label{tab:ncapfull}
\end{table*}

\subsection{The speed floor}
\label{sec:nofloor}
\begin{table}[!t]
\centering\scriptsize\setlength{\tabcolsep}{4pt}
\resizebox{\columnwidth}{!}{\begin{tabular}{@{}llrr@{}}
\toprule
metric & \model & without floor & with floor (Eq.~\ref{eq:unicycle}) \\
\midrule
Table~\ref{tab:traj} mean$_5$ (m) & speed head, zero-shot & 2.545 & 2.532 \\
Table~\ref{tab:traj} mean$_5$ (m) & speed head, few-shot & 1.490 & 1.476 \\
Table~\ref{tab:traj} mean$_5$ (m) & speed sensor & 1.120 & 1.108 \\
Table~\ref{tab:traj} mean$_5$ (m) & speed sensor + ego history & 1.111 & 1.098 \\
\midrule
Table~\ref{tab:traj} top-1 (m) & speed head, zero-shot & 5.078 & 5.064 \\
Table~\ref{tab:traj} top-1 (m) & speed head, few-shot & 3.600 & 3.573 \\
Table~\ref{tab:traj} top-1 (m) & speed sensor & 3.012 & 2.971 \\
Table~\ref{tab:traj} top-1 (m) & speed sensor + ego history & 2.994 & 2.952 \\
\midrule
Euro NCAP score & 1 front, 75$^\circ$ & 0.291 & 0.296 \\
Euro NCAP score & 3 front, 150$^\circ$ & 0.362 & 0.365 \\
Euro NCAP pass rate (\%) & 1 front, 75$^\circ$ & 14.0 & 15.4 \\
Euro NCAP pass rate (\%) & 3 front, 150$^\circ$ & 18.3 & 19.1 \\
\midrule
DCA-Bench pass rate (\%) & 1 front, 75$^\circ$ & 39.3 & 39.3 \\
\bottomrule
\end{tabular}}
\caption{\textbf{Without and with the speed floor of Equation~\ref{eq:unicycle}.} The same weights, inputs and selection; only the roll-out of the chosen controls differs. Trajectory: the mean$_5$ and top-1 columns of Table~\ref{tab:traj}, both arms run in the same environment on the frames resized without the framing of Section~\ref{sec:framing}; the comparison has not been repeated on the framed clips, so the with-floor column is not Table~\ref{tab:traj}'s, and the few-shot rows of both arms were fitted with the earlier calibration rule (Section~\ref{sec:calibration}). Euro NCAP: the score and pass rate of Table~\ref{tab:ncap}, both from a same-machine redump. DCA-Bench: the pass rate over the 89 incident--twin pairs, where every verdict is the same with and without the floor.}
\label{tab:nofloor}
\end{table}

Table~\ref{tab:nofloor} gives the main \model results with the candidates rolled out without the floor at zero speed. Every pick is the same in both columns; without the floor, a candidate that brakes to a stop rolls backward along its path.

\subsection{DCA-Bench hazard-free twins}
\begin{table}[!htbp]
\centering\scriptsize\setlength{\tabcolsep}{3pt}
\resizebox{\columnwidth}{!}{\begin{tabular}{@{}lrrrr@{}}
\toprule
 & \multicolumn{2}{c}{incidents passed (\%)} & \multicolumn{2}{c}{twins} \\
\cmidrule(lr){2-3}\cmidrule(lr){4-5}
model & published & drivable & FP rate (\%) & pairs passed (\%) \\
\midrule
Full brake, 6 m/s$^2$ (control) & 98.0 & 98.0 & 93.3 & 6.7 \\
Recorded driver & 45.0 & 45.0 & 0.0 & 44.9 \\
\midrule
\textbf{\model} & 56.0 & 46.0 & 13.5 & 39.3 \\
\model, head's own pick & 32.0 & 25.0 & 13.5 & 20.2 \\
Alpamayo-R1$^\dagger$ & 19.5 & 19.5 & 9.7 & 18.2 \\
Alpamayo~2~Super$^\dagger$ & 19.5 & 19.5 & 13.1 & 14.2 \\
Qwen-Drive SFT & 11.0 & 11.0 & 24.7 & 7.9 \\
Alpamayo~1.5$^\dagger$ & 12.7 & 12.7 & 18.2 & 9.9 \\
Qwen-Drive RL & 14.0 & 14.0 & 73.0 & 6.7 \\
SimWAM SFT & 10.0 & 4.0 & 85.4 & 1.1 \\
SimWAM RL & 9.0 & 4.0 & 92.1 & 0.0 \\
\bottomrule
\end{tabular}}
\caption{\textbf{DCA-Bench in full.} Incidents passed by the contact check only, and after the drivability check (no more than 8.8 m/s$^2$ of braking from the true speed), both of 100; false positives on the twins and pairs passed, both of the 89 pairs. $^\dagger$Mean of 6 runs, each paired with the same run on its twin.}
\label{tab:dcafull}
\end{table}
Each twin is cut from its incident's clip so that the model's input and the whole 4 s horizon end before the challenge's alert time, the first moment the hazard is visible; other traffic is allowed. Eleven incidents have no twin: on three the recorded path touches a road user before the alert, on two the driver already brakes above 3.5 m/s$^2$, one reconstruction starts too late, and on five the twin would start before the public challenge clip does. Every planner reads the public challenge clip, for the incident and the twin, with the inputs of its incident run; the twin's frames are placed in the clip by content and released with the benchmark. A plan's speed is read over 0.2 s, starting from the car's true speed, so a plan that starts below the true speed counts as braking; the same rule gives the drivability check on the incidents. The external planners' ego history comes from the reconstruction, which is smoothed over the whole clip. Table~\ref{tab:dcafull} gives every count.

\subsection{What one pass produces on unseen roads}
\begin{figure*}[!t]
  \centering
  \includegraphics[width=0.78\textwidth]{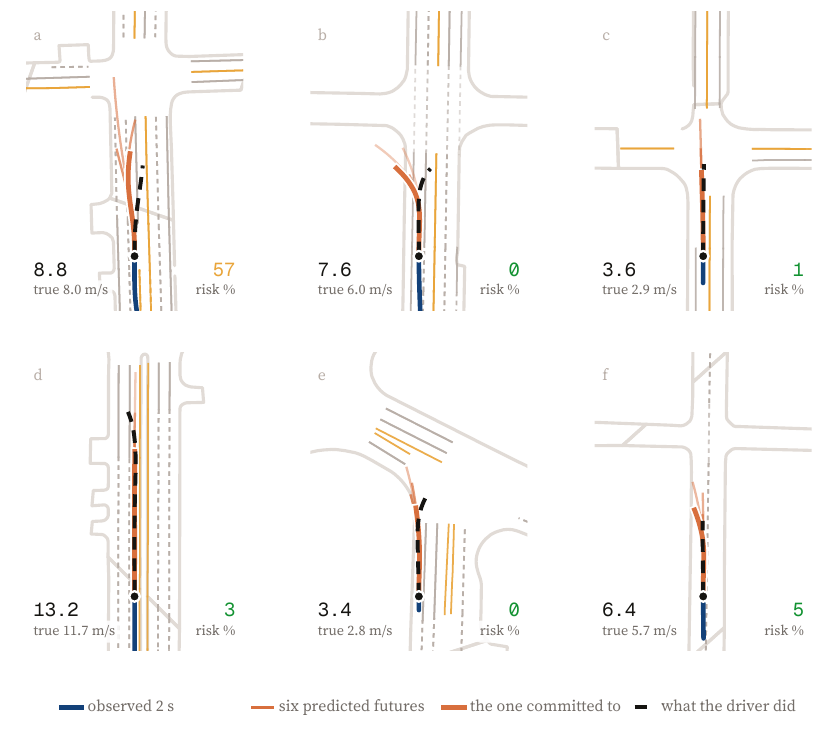}
  \caption{\textbf{Six Argoverse~2 windows at one scale}: the observed two seconds in blue, six futures in light orange with the committed one (the safety selector's pick) drawn heavy, the driven path dashed in black, and the speed and risk from the same pass. Road geometry is the AV2 vector map, drawn for the reader; the model sees no map and no intrinsics.}
  \label{fig:uncertainty}
\end{figure*}
Figure~\ref{fig:uncertainty} shows six Argoverse~2 windows from one pass each: the futures spread where the road allows and close where it does not, with speed and risk read alongside.

\subsection{Backbone study, all four tasks}
\label{sec:backbone}
\begin{figure*}[!t]
  \centering
  \includegraphics[width=0.8\textwidth]{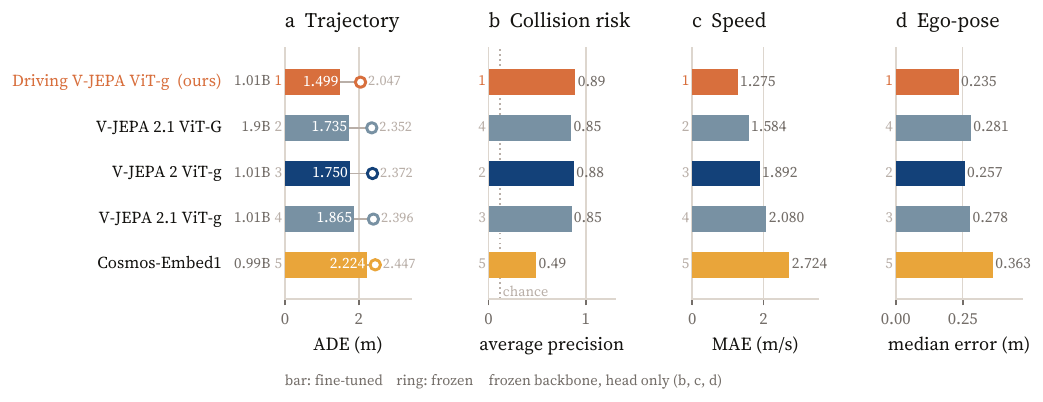}
  \caption{\textbf{Five encoders on four tasks.} (a) Trajectory validation ADE, fine-tuned bars with the frozen depth-mix readout as a marker; (b) risk AP, (c) speed error, (d) ego-pose error, each a dedicated head on the frozen representation.}
  \label{fig:backbone}
\end{figure*}

We compare our chosen backbone with four different pretrained backbones across each of the four outputs \model produces; Meta's V-JEPA~2~\cite{vjepa2} ViT-g and V-JEPA~2.1~\cite{vjepa21} at ViT-g and the 1.9B ViT-G, and NVIDIA's Cosmos-Embed1~\cite{cosmosembed1}. This compares V-JEPA models trained on in-domain data (the backbone of \model) vs general video data. Cosmos-Embed1 is a joint video-text embedder built for physical AI. With each backbone we train dedicated heads for trajectory forecasting, collision anticipation, ego-pose estimation and speed estimation, all with the backbones frozen. We also train a trajectory forecasting head but with each backbone trainable. The results show that the domain-trained V-JEPA backbone is superior on all tasks bar last-layer trajectory forecasting (Table~\ref{tab:backbonefull}).\looseness=-1 Table~\ref{tab:backbonefull} and Figure~\ref{fig:backbone} give the comparison on all four tasks.
\begin{table*}[!t]
\centering\scriptsize\setlength{\tabcolsep}{4pt}
\begin{tabular}{@{}lrrrrrrr@{}}
\toprule
backbone & params & frozen, last layer & frozen, four depths & fine-tuned & risk AP & speed MAE & ego-pose ATE \\
\midrule
\textbf{Driving V-JEPA ViT-g} & 1.01B & 2.198 & 2.047 & 1.499 & 0.893 & 1.27 & 0.235 \\
V-JEPA 2.1 ViT-G & 1.90B & 2.273 & 2.352 & 1.735 & 0.848 & 1.58 & 0.281 \\
V-JEPA 2 ViT-g & 1.01B & 2.029 & 2.372 & 1.750 & 0.877 & 1.89 & 0.257 \\
V-JEPA 2.1 ViT-g & 1.01B & 2.522 & 2.396 & 1.865 & 0.853 & 2.08 & 0.278 \\
Cosmos-Embed1 & 0.99B & 2.355 & 2.447 & 2.224 & 0.490 & 2.72 & 0.363 \\
\bottomrule
\end{tabular}
\caption{\textbf{The backbone study in one table.} Trajectory validation ADE (m) with the backbone frozen and read at the last layer, frozen and read across four depths, and fine-tuned; then the frozen-probe risk AP against the teacher, speed MAE (m/s) and ego-pose median ATE (m). All values are on validation windows.}
\label{tab:backbonefull}
\end{table*}

\subsection{Runtime}
\begin{table}[!htbp]
\centering\scriptsize\setlength{\tabcolsep}{4pt}
\resizebox{\columnwidth}{!}{\begin{tabular}{@{}lrrlrr@{}}
\toprule
model & params & input & GPU & median latency & peak memory \\
\midrule
\textbf{BADAS-3.0}, compiled & 1B & 1 cam & A100 & 43 ms & 1.9 GiB \\
\textbf{BADAS-3.0}, eager & 1B & 1 cam & A100 & 146 ms & 2.0 GiB \\
Qwen-Drive 4B, direct & 4B & 3 cam & A100 & 967 ms & 12.3 GiB \\
Qwen-Drive 4B, reasoning & 4B & 3 cam & A100 & 2.0 s & 11.2 GiB \\
Alpamayo-R1 & 10B & 4 cam & A100 & 4.1 s & 25.9 GiB \\
Alpamayo 1.5 & 10.5B & 4 cam & A100 & 4.4 s & 26.1 GiB \\
Alpamayo 2 Super & 34B & 6 cam & A100 & 4.9 s & 77.5 GiB \\
SimWAM & -- & 1 cam & A100 & 10.6 s & 12.9 GiB \\
\midrule
\textbf{BADAS-3.0}, compiled & 1B & 1 cam & B200 & 57 ms & 2.0 GiB \\
Alpamayo 2 Edge, release acceleration & 2B & 4 cam & B200 & 223 ms & 7.4 GiB \\
\bottomrule
\end{tabular}}
\caption{\textbf{Latency and peak memory per two-second window}, batch one, bf16, median of 30 calls after warm-up, on the GPU named in each row (one A100-80GB or one B200). Alpamayo~2~Edge was timed on a B200 with its release's own acceleration, so \model is timed on that machine as well. The \model rows time the backbone and the trajectory head, which emits all six candidates; the risk head and the safety selector add 1.5 ms, and the ego-pose and speed heads under 1 ms.}
\label{tab:runtime}
\end{table}
\begin{figure}[!htbp]
  \centering
  \includegraphics[width=\columnwidth]{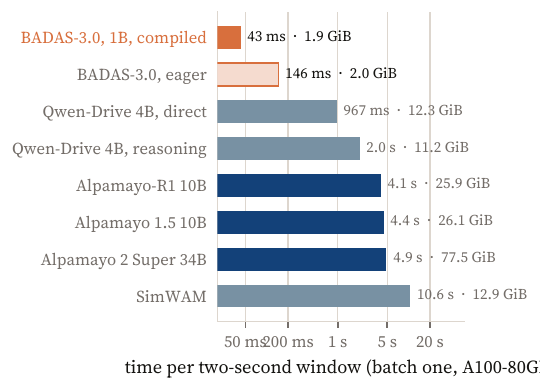}
  \caption{\textbf{Time per two-second window}, batch one, each model's own release code; peak memory beside each bar. One A100-80GB, except the two rows marked B200.}
  \label{fig:runtime}
\end{figure}
The compiled \model path moves waypoints by 0.84\% relative to eager; the accuracy tables use the eager path.

\subsection{Head architectures}
\label{sec:headarch}
All four heads read the 2048 patch tokens of 1408 dimensions produced by the backbone.

\textbf{Trajectory head.} The given ego history is embedded into 20 tokens, which are combined with the patch tokens to form the memory tokens. If no ego history is provided, learned null tokens take their place. Six learned mode query tokens pass through three layers. In each layer the queries first attend to each other by self-attention, then to the memory tokens by cross-attention, and finally go through a feed-forward network. This gives one hidden state $\mathbf{h}_k \in \mathbb{R}^d$ per mode. A single linear layer projects each $\mathbf{h}_k$ to a vector of dimension 128, which is reshaped to $64\times2$ to give the deltas $\Delta_k$.

\textbf{Ego-pose head.} The patch features are mean-pooled over $4\times4$ spatial blocks, which leaves 128 tokens ($4\times4\times8$), and a linear layer projects each token from 1408 to 256 dimensions. A single query token attends to these tokens by cross-attention and is passed through an MLP, with a final projection to the four output parameters $[a_0, a_1, \kappa_0, \kappa_1]$.

\textbf{Speed head.} The speed head has the same architecture as the ego-pose head, with a final projection to a single scalar. The scalar is interpreted as $\log(\hat{v}+0.1)$ and is trained with a Huber loss against $\log(v+0.1)$.

\textbf{Risk head.} A single query token attends to all 2048 patch tokens by cross-attention, and an MLP maps the result to the output logit.

\subsection{Training details}
Trajectory loss: Huber ($\beta{=}1$ m) on the integrated $xy$ of the closest of six modes plus 0.5 cross-entropy on its index; the unicycle inside the loss advances the pose with the current state and then applies the current control. Risk: binary cross-entropy, weight 5, to the teacher's probability ($\lambda=0.5$ then $0.9$ in the joint stage with the remainder on the hard collision label; $\lambda=1$ in the head-only continuation), over 1.88M windows. Speed: Huber on $\log(v_0+0.1)$. Ego-pose: Huber on the twenty past points. Stitcher: Huber on yaw and $xy$ of the fused track. The backbone is trained in the trajectory and joint stages and frozen afterwards, when the remaining heads are trained on cached tokens of 243,720 windows. Compute for the final weights: about 2,700 GPU-hours of pretraining and about 600 of adaptation (about 400 on eight A100s for the trajectory stage, 190 on eight H100s for the joint stage and 12 for the frozen-backbone stages). Parameters measured from the released checkpoint: 1,012.2M backbone and 17.8M heads.

\subsection{Joint risk training}
\label{sec:control}
In the joint stage of Section~\ref{sec:training} the risk head is trained from initialisation while the trajectory head is already partly trained, and the backbone receives gradients from both objectives on different batches. During this stage the validation minADE$_6$ improves from 1.225 m to 1.209 m; a run from the same initialisation for the same number of steps without the risk objective reaches 1.208 m, so joint training does not hinder trajectory performance.\looseness=-1

\subsection{NavSim details}
\begin{table}[!t]
\centering\scriptsize\setlength{\tabcolsep}{3pt}
\resizebox{\columnwidth}{!}{\begin{tabular}{@{}lll@{}}
\toprule
method & sensors & PDMS \\
\midrule
\multicolumn{3}{@{}l}{\emph{no reinforcement learning}} \\
TransFuser~\cite{transfuser2023} & 3 cam + LiDAR & 84.0 \\
Epona~\cite{epona2025}, world model & 1 cam & 86.2 \\
DiffusionDrive~\cite{diffusiondrive} & 3 cam + LiDAR & 88.1 \\
Qwen-Drive-1.0~\cite{qwendrive}, SFT, reasoning & 3 cam & 88.2 \\
BADAS-3.0, recalibrated & 1 cam & 89.0 \\
BADAS-3.0, LoRA & 1 cam & 90.0 \\
\textbf{SimWAM}~\cite{simwam}, SFT & 1 cam & \textbf{90.3} \\
\midrule
\multicolumn{3}{@{}l}{\emph{reinforcement learning}} \\
AutoVLA~\cite{autovla2025} & 3 cam & 89.1 \\
ExploreVLA~\cite{explorevla2026} & 1 cam & 90.4 \\
Qwen-Drive-1.0~\cite{qwendrive} & 3 cam & 90.7 \\
ReCogDrive~\cite{recogdrive2026} & 1 cam & 90.8 \\
ReflectDrive-2~\cite{reflectdrive22026} & 3 cam & 91.0 \\
BADAS-3.0, LoRA + GRPO & 1 cam & 91.1 \\
\textbf{SimWAM}~\cite{simwam} & 1 cam & \textbf{91.9} \\
\bottomrule
\end{tabular}}
\caption{\textbf{NavSim v1 navtest, PDM score of one committed trajectory}, 12,146 scenes. Published rows are copied from their papers and were not re-run. The \model rows commit the candidate a four-seed selector ensemble picks from the six (three seeds for the GRPO row): with the trajectory head's controls recalibrated, after a LoRA fine-tune of backbone and head on navtrain, and after end-to-end GRPO through the LoRA backbone, head and a co-trained selector against the PDM score. The supplement gives the NavSim protocol in detail.}
\label{tab:navsim}
\end{table}

\begin{table}[!htbp]
\centering\scriptsize\setlength{\tabcolsep}{4pt}
\resizebox{\columnwidth}{!}{\begin{tabular}{@{}lrr@{}}
\toprule
method & $N$ & best-of-$N$ PDMS \\
\midrule
Qwen-Drive-1.0~\cite{qwendrive}, SFT & 6 & 89.3 \\
Qwen-Drive-1.0~\cite{qwendrive}, RL & 6 & 91.4 \\
AutoVLA~\cite{autovla2025} & 6 & 92.1 \\
DriveVLA-W0~\cite{drivevlaw02026} & 6 & 93.0 \\
AdaThinkDrive~\cite{adathinkdrive2025} & 4 & 93.0 \\
ExploreVLA~\cite{explorevla2026} & 6 & 93.7 \\
Curious-VLA~\cite{curiousvla2026} & 6 & 94.8 \\
ReflectDrive-2~\cite{reflectdrive22026} & 6 & 94.8 \\
\midrule
BADAS-3.0 & 6 & 86.4 \\
BADAS-3.0, recalibrated & 6 & 95.7 \\
BADAS-3.0, LoRA & 6 & 95.5 \\
\textbf{BADAS-3.0, LoRA + GRPO} & 6 & \textbf{96.3} \\
\bottomrule
\end{tabular}}
\caption{\textbf{NavSim v1 navtest, best-of-$N$ PDMS.} The best candidate per scene, chosen with the score, so these are ceilings of a candidate set and not planner results. Published rows sample $N$ trajectories and are copied from their papers. \model rows are the best of its six anchored candidates, from the released checkpoint, after recalibrating the controls, after the LoRA fine-tune and after end-to-end GRPO.}
\label{tab:navsimbestofn}
\end{table}

\textbf{Training stages.} The recalibration and the LoRA fine-tune each start from the standard \model, and the GRPO stage continues from the LoRA model. The NavSim runs predate the floor at zero in Equation~\ref{eq:unicycle}: their candidates are rolled out without it, and only the trajectories of candidates that come to a stop would differ. Under best-of-$6$ evaluation, \model scores $86.4$ on navtest.

\emph{Recalibration.} The backbone and all heads stay frozen. Six calibration parameters are applied to the trajectory head's controls and fitted against 4,000 scenes from navtrain using the human trajectory as ground truth. The six parameters are made up of a scale, shift and ramp coefficient applied to each of the acceleration and curvature channels: $x \mapsto ax + b + ct$ where $t$ ranges from $0$ to $1$ over the trajectory. This raises the best-of-$6$ performance to $95.7$.

\emph{LoRA fine-tune.} We add rank-4 LoRA adapters to the attention and MLP layers of all 40 backbone layers and train them together with the trajectory head, 7.8M trainable parameters in all. The objective is the imitation objective of the original trajectory training (Section~\ref{sec:training}), run for two passes over the full navtrain training scenes. This achieves $95.5$ under best-of-$6$.

\textbf{Selector.} To obtain a single predicted trajectory from the model's candidates we fit a selector module. This predicts the PDM score of each candidate from which we take the argmax as our selection. The selector is a single self-attention block over the six candidate trajectories and trained with a squared-error loss against the true PDM scores. We fit a selector over four random seeds (three for the GRPO model) and ensemble them to form our final module. Using this selector architecture the recalibrated model achieves 89.0 on navtest and the LoRA model 90.0.

\emph{GRPO.} Starting from the LoRA model we continue with direct GRPO to jointly optimise the model and the selector. The reward is the PDM score of a sampled trajectory, and the LoRA adapters, the trajectory head and the selector are trained together. An imitation term and a KL penalty keep the model close to where it started. We train for 16 passes over the 16,000 navtrain scenes whose candidates include a gate-failing mode. This model achieves 96.3 under best-of-$6$ and 91.1 with the selector on navtest.

We also tested a braking offset, one scalar added to the recalibrated acceleration bias and swept against the true PDM scorer on 640 navtrain scenes. With it the selected score is 88.97, without it 88.98 (paired difference without minus with $+0.004$, log-cluster interval $[-0.47, +0.44]$), so we report the version without it, in which the scorer never touches the model. A candidate set's ceiling does not predict the selected score either. The recalibrated controls without the offset have the highest oracle (95.7) and tie the version with it on the selected score. The recalibration needs few scenes when measured in held-out displacement error, where eight scenes recover about 90\% of the improvement, but displacement error and PDM score do not track each other on this model, so we do not claim the same for PDM score.

\subsection{Preparing a camera: input framing and calibration}
\label{sec:framing}
A camera is prepared in two steps. Both are fixed once per camera and then reused for all of its video, and neither changes the weights. The first, framing, uses only the camera's known parameters and needs no labels; the zero-shot rows use it alone. The second, calibration, fits one scale per output from ten labelled windows, and is what the few-shot rows add.

\textbf{Framing.} \model was trained on 16:9 cameras whose focal length is 0.48 (PhysicalAI-AV front wide) to 0.60 (the dashcam) of the frame width. The front cameras of nuScenes (0.79), Argoverse~2 (1.09 to 1.15) and Waymo (1.08) are narrower. For these sets each native frame is placed, before the $256\times256$ resize, on a 16:9 canvas whose width is the focal length divided by 0.60, with black bars where the canvas is larger than the frame and a centre crop where the frame is taller. The target, 0.60, is the training dashcam's measured value and the focal lengths come from each dataset's calibration; no parameter is fitted to benchmark results. KITTI, PAI-644, PAI-100c and the dashcam sets are inside the training range and are resized as they are. Windows, ground truth, weights and every other method's inputs are the same with and without the framing. Against resizing the full frame, the framing lowers \model's error on these sets in every cell of Tables~\ref{tab:traj} and~\ref{tab:ego} and in every zero-shot cell of Table~\ref{tab:speed}. Three results get worse: the few-shot speed error on Argoverse~2 Sensor, compared under the earlier calibration rule (0.46 to 0.47 m/s; 0.46 m/s with Equation~\ref{eq:gate}); the shape error after the similarity fit on Argoverse~2 Sensor (2.5 to 3.1 cm zero-shot) and Waymo (3.8 to 4.0 cm, Table~\ref{tab:egosim3}); and the two-second error when the speed is given by the sensor, a reference that is not in Table~\ref{tab:ego} (0.14 to 0.21 m on Argoverse~2 Sensor and 0.15 to 0.19 m on TbV). The AV2 and Waymo segments of the full drives (Table~\ref{tab:full}) are framed the same way, frame by frame at their native resolution. There the framing lowers every zero-shot and sensor-speed cell without alignment and every cell after the similarity fit, and, compared under the earlier calibration rule, raised the few-shot error on AV2 (3.09 to 5.38 m stitched, 2.32 to 3.78 m chained); with Equation~\ref{eq:gate} these two cells are 2.37 and 2.73 m.

\textbf{Calibration.}\label{sec:calibration} The first ten usable labelled windows of a set in id order (segments in Table~\ref{tab:full}) are set aside and are not scored. Each gives a pair of a predicted quantity $\hat y_i$ and its true value $y_i$. For the trajectory and speed rows these are the speed head's output and the sensor speed at $t_0$; for the ego-pose rows they are the distance per second along the row's own zero-shot track and along the true path. A window is usable when the true value is above 1 m/s and the predicted one above 0.1 m/s; the speed rows of Table~\ref{tab:speed} set aside exactly a rig's first ten windows and fit on the usable ones among them. With $n=10$ pairs,
\begin{equation}
\hat k=\frac{\sum_i y_i}{\sum_i \hat y_i},\qquad
\mathrm{se}=\frac{\sqrt{\frac{n}{n-1}\sum_i (y_i-\hat k\,\hat y_i)^2}}{\sum_i \hat y_i},
\label{eq:khat}
\end{equation}
\begin{equation}
k=\begin{cases}1 & \text{if } |\hat k-1|\le\max(2\,\mathrm{se},\,0.05),\\[2pt] \dfrac{n\hat k+1}{n+1} & \text{otherwise.}\end{cases}
\label{eq:gate}
\end{equation}
The speed head's output is multiplied by $k$ before the candidates and the pose track are integrated; the predicted controls and the risk are untouched. The ratio of sums weights a window by its speed. The first case of Equation~\ref{eq:gate} is a gate: the zero-shot prediction is kept when ten windows cannot tell $k$ from one, or when the factor is within 5\% of one. The second case adds one pseudo-observation at $k=1$. Where the gate is closed, the few-shot row is the zero-shot prediction scored without the calibration windows, so it differs from the zero-shot cell only through those windows. In Table~\ref{tab:traj} the gate is closed on nuScenes ($\hat k=1.02$, $\mathrm{se}=0.05$) and PAI-644 ($\hat k=0.99$, $\mathrm{se}=0.03$) and open on Argoverse~2 Sensor ($k=0.89$), TbV ($k=0.91$) and Waymo ($k=1.28$). In Tables~\ref{tab:ego}, \ref{tab:speed} and~\ref{tab:full} it is closed on nuScenes (speed and ego-pose) and for the stitched row on AV2, and open on every other set, with $k$ from 0.87 (Argoverse~2 Sensor, ego-pose) to 1.45 (KITTI, ego-pose).

Earlier versions of these tables used an earlier rule, the shrunk geometric mean of the ten per-window ratios, applied on every set. Against it, the few-shot row of Table~\ref{tab:traj} moves from 1.19 to 1.22 m on nuScenes, where it gets worse, from 1.31 to 1.23 m on Argoverse~2 Sensor and from 1.08 to 1.06 m on TbV, and by less than 0.01 m on Waymo and PAI-644; its mean falls from 1.27 to 1.26 m and its top-1 mean rises from 3.28 to 3.30 m. The few-shot speed error of Table~\ref{tab:speed} gets worse on nuScenes, TbV, KITTI (by 0.002 m/s) and PAI-100c (0.45 to 0.52 m/s) and its mean rises from 0.51 to 0.52 m/s; the few-shot ego-pose mean of Table~\ref{tab:ego} rises from 0.396 to 0.400 m. In Table~\ref{tab:full} the AV2 cells fall from 5.38 to 2.37 m stitched and from 3.78 to 2.73 m chained, and the Waymo cells rise from 4.16 to 4.70 m and from 4.03 to 4.15 m. The PAI-644 few-shot cells of Tables~\ref{tab:ego} and~\ref{tab:full} keep the earlier rule ($^{\ddagger}$), because they were scored from aggregates of a cut that can no longer be rebuilt. The proprietary cells of Table~\ref{tab:full} also keep it ($^{\ddagger}$); they were not re-run.

\textbf{Held-out check.} The rule (the ratio of sums, the gate at two standard errors, the pseudo-observation, and distance pairs for the ego-pose rows) was selected using the benchmark sets themselves, which is model selection on the reported sets. It was then frozen and checked once on four ego-pose cells that took no part in the selection: the two-second windows of KITTI, and the two-second windows and 15-second segments of PAI-100c. The criteria were written down beforehand: not worse than zero-shot by more than 2\%, and at least as good as the earlier rule on most cells. Table~\ref{tab:heldout} gives the result.

\begin{table*}[!t]
\centering\footnotesize\setlength{\tabcolsep}{5pt}
{\begin{tabular}{@{}lrrrrlll@{}}
\toprule
held-out cell & $n$ & zero-shot & earlier rule & frozen rule & ($\hat k$, se) & result & with the 5\% minimum \\
\midrule
KITTI, 2 s & 102 & 3.059 & 0.525 & 0.538 & (1.491, 0.041) & pass & 0.538 (unchanged) \\
PAI-100c, 2 s & 187 & 0.312 & 0.327 & 0.367 & (1.062, 0.012) & fail, $+17.5\%$ & 0.367 (still applied) \\
PAI-100c, 15 s stitched & 90 & 2.361 & 2.501 & 2.504 & (1.046, 0.014) & fail, $+6.1\%$ & zero-shot (gate closed) \\
PAI-100c, 15 s chained & 90 & 2.656 & 2.836 & 2.582 & (1.052, 0.015) & pass, $-2.8\%$ & 2.582 (still applied) \\
\bottomrule
\end{tabular}}
\caption{\textbf{Held-out check of the calibration rule.} Median ATE in metres with no alignment, every column on the same scored windows. Frozen rule: Equation~\ref{eq:gate} with the gate at two standard errors only. Result: the frozen rule against zero-shot. The last column adds the 5\% minimum; it was introduced after the check and is not an independent result.}
\label{tab:heldout}
\end{table*}

The check is mixed. Two of the four cells pass, and the frozen rule is not better than the earlier rule on most cells. The earlier rule is also worse than zero-shot on all three PAI-100c cells. On that camera the offset is real but small, about 3.5\%, and ten windows cannot resolve it ($\hat k$ between 1.00 and 1.07 over random draws). The 5\% minimum in Equation~\ref{eq:gate} was added after this check and because of it, so the last column of Table~\ref{tab:heldout} is not an independent result. It closes the gate on one cell, the stitched 15-second cell; on the chained one $\hat k=1.052$ is just above the minimum. The check ran in fp32. The published KITTI and PAI-100c few-shot cells are bf16 runs, the precision of their zero-shot rows, so they differ from these by a few percent (KITTI, 2 s: 0.532 m). What the evidence supports is narrower than a general claim for the rule: ten labelled windows correct a gross scale error (KITTI, Waymo and Argoverse~2, 43 to 83\% lower speed and ego-pose error under either rule), and for a camera inside the training range the zero-shot row is the one to use.

\subsection{Reproducibility details}
Frames are resized to $256\times256$ with area interpolation and normalised with ImageNet statistics. The speed head predicts $\log(v_0+0.1)$ and is read as $\exp(\cdot)-0.1$ clamped at zero, so standstill is representable. During trajectory training the roll-out speed is the sensor speed, the displacement between the last two present poses over 0.1 s. The risk label is the teacher's window probability. Anchors: $k$-means with $K{=}6$ over the training $(a,\kappa)$ sequences, one real member per cluster, stored as a buffer; the delta layer is zero-initialised. Closest-mode loss: Huber ($\beta{=}1$ m) on the integrated $xy$ of the closest mode plus 0.5 cross-entropy on its index. The stitcher fuses consecutive per-window controls by inverse-variance weighting into one 10 Hz track and is trained with Huber losses on yaw and $xy$ of the fused track. Adapter sign conventions were checked on sequences of known turn direction. The parameter count of the released checkpoint is 1,012.2M for the backbone and 17.8M for the four heads.

\end{document}